\documentclass[letterpaper, 10pt, conference]{ieeeconf}   

\IEEEoverridecommandlockouts               
\usepackage{cite}
\usepackage{color,xcolor}
\usepackage{graphicx}
\usepackage{pdfpages}
\usepackage{subfigure}
\usepackage{tabulary}

\usepackage{amsmath}
\usepackage{amssymb}
\usepackage{float}
\usepackage{soul}
\usepackage{booktabs} 
\usepackage{multirow}  
\usepackage{hyperref} 
\usepackage{array}  
\usepackage{diagbox}

\usepackage{bm}
\usepackage{verbatim}

\begin{document}

\title{Nonlinear Model Predictive Control for Robust Bipedal Locomotion: Exploring Angular Momentum and CoM Height Changes}

\author{Jiatao Ding\textsuperscript{a,b},
        Chengxu Zhou\textsuperscript{c}, Songyan Xin\textsuperscript{d}, Xiaohui Xiao\textsuperscript{a}$^\ast$ and Nikos Tsagarakis\textsuperscript{e}
\thanks{\textsuperscript{a}School of Power and Mechanical Engineering, Wuhan University, Wuhan, China.}
\thanks{\textsuperscript{b}Shenzhen Institute of Artificial Intelligence and Robotics for Society, ShenZhen, China.}
\thanks{\textsuperscript{c}School of Mechanical Engineering, University of Leeds, Leeds, UK.}
\thanks{\textsuperscript{d}School of Informatics, The University of Edinburgh, Edinburgh, UK.}
\thanks{\textsuperscript{e}Humanoid and Human Centered Mechatronics Research Line, Istituto Italiano di Tecnologia, Genoa, Italy}
\thanks{\textit{Corresponding Author}: xhxiao@whu.edu.cn}
}
\maketitle

\begin{abstract}
Human beings can utilize multiple balance strategies, e.g. step location adjustment and angular momentum adaptation, to maintain balance when walking under dynamic disturbances. In this work, we propose a novel Nonlinear Model Predictive Control (NMPC) framework for robust locomotion, with the capabilities of step location adjustment, Center of Mass (CoM) height variation, and angular momentum adaptation. These features are realized by constraining the Zero Moment Point within the support polygon. By using the nonlinear inverted pendulum plus flywheel model, the effects of upper-body rotation and vertical height motion are considered. As a result, the NMPC is formulated as a quadratically constrained quadratic program problem, which is solved fast by sequential quadratic programming. Using this unified framework, robust walking patterns that exploit reactive stepping, body inclination, and CoM height variation are generated based on the state estimation. The adaptability for bipedal walking in multiple scenarios has been demonstrated through simulation studies.
\end{abstract}

\IEEEpeerreviewmaketitle

\section{Introduction}
Humanoid robots have attracted much attention for their capabilities in accomplishing challenging tasks in real-world environments. With several decades passed, state-of-the-art {robot platforms} such as ASIMO \cite{sakagami2002intelligent}, Atlas \cite{fallon2015architecture}, WALK-MAN \cite{tsagarakis2017walk}, and CogIMon \cite{zhou2018ik} have been developed for this purpose. However, due to the complex nonlinear dynamics of bipedal locomotion over the walking process, enhancing walking stability, which is among the prerequisites in making humanoids practical, still needs further studies. In this paper, inspired by the fact that human beings can make use of the redundant Degree of Freedom (DoF) and adopt various strategies, such as the ankle, hip, and stepping strategies, to realize balance recovery \cite{horak1986central,nashner1985organization,orendurff2004effect}, {we aim to develop a versatile and robust walking pattern generator which can integrate multiple balance strategies in a unified way}.

To generate the walking pattern in a time-efficient manner, simplified dynamic models have been proposed, among which the Linear Inverted Pendulum Model (LIPM) is widely used \cite{kajita1991study}. Using the LIPM, Kajita et al. proposed the preview control for Zero Moment Point (ZMP) tracking \cite{kajita2003biped}. By adopting a Linear Quadratic Regulator (LQR) scheme, the ankle torque was adjusted to modulate the ZMP trajectory and Center of Mass (CoM) trajectory. As a result, robust walking can be achieved in the presence of external perturbations. Nevertheless, this strategy can neither modulate the step parameters nor take into consideration the feasibility constraints arisen from actuation limitations and environmental constraints. To overcome this drawback, Wieber et al. proposed a Model Predictive Control (MPC) algorithm to utilize the ankle strategy \cite{wieber2006trajectory} and then extended it for adjusting step location \cite{diedam2008online}. Since then, by utilizing the stepping strategy (step location adjustment), stable walking on the uneven surface, unknown slope, and balance recovery from external pushes have been realized \cite{feng2016robust,ding2018walking,castano2016dynamic,kamioka2017dynamic}. However, the absence of angular momentum adaptation and vertical height variation limits the capability against dynamic disturbances.

The angular momentum adaptation, which is also called as the hip strategy, plays a vital role in enhancing walking robustness against external disturbances \cite{horak1986central}.  Based on the general ZMP dynamics which took into account the moment around the CoM, a Quadratic Programming (QP) algorithm was employed in \cite{kudoh2002dynamic} to determine the body inclination state. Then, by using the simplified Linear Inverted Pendulum plus Flywheel Model (LIPFM), the roll and pitch angular momentum\footnote{In some studies such as \cite{aftab2012ankle,lack2015integrating}, the roll and pitch angular momentum of upper-body around the CoM are also referred as spin angular momentum.} adaptation were realized by rotating the upper-body \cite{pratt2006capture}.  Focusing on the hip strategy, Li et al. proposed an open-loop control algorithm for rejecting the external pushes \cite{li2015push}. Furthermore, by employing the LIPFM, Aftab et al. adopted a Nonlinear MPC (NMPC) framework \cite{aftab2012ankle}, and Jeong et al. proposed a Quadratic Programming (QP) scheme \cite{jeong2019robust}, which can both integrate the ankle, stepping, and hip strategies. Nevertheless, since the LIPFM still assumes a constant CoM height plane,  above works could not make use of the height variation. 

Vertical height motion also contributes to higher walking robustness \cite{horak1986central,massaad2007up}. Nishiwaki et al. proposed a trajectory planning algorithm, which could generate time-varying height trajectories \cite{nishiwaki2011online}. However, this work did not consider the variation of the ZMP dynamics. The same problem also exists in \cite{mirjalili2018whole}. Englsberger et al. \cite{englsberger2015three} solved the vertical height motion through generalizing the 3D divergent component of motion, whereas it could not consider feasibility constraints. In term of ZMP dynamics, the nonlinear motion equations of the general inverted pendulum model that were derived in \cite{kudoh2002dynamic,kajita2007human,wieber2016modeling} can be used to model the height variation.  Through limiting the nonlinear part of the ZMP motion constraint between properly chosen extreme values, Brasseur et al. built a Linear MPC (LMPC) scheme and realized the real-time generation of 3D walking gait \cite{brasseur2015robust}. Van Heerden solved the NMPC problem efficiently via Sequential Quadratic Programming (SQP) after formulating the optimization problem as a Quadratically Constrained Quadratic Program (QCQP) problem \cite{van2017real}. Then, by introducing the concept of \textit{stiffness},  Caron et al. studied the dynamics property of the so-called Variable-Height Inverted Pendulum (VHIP) in term of capturability analysis. Consequently, the nonlinear feasibility constraints were linearized, and a hierarchical optimization scheme was proposed for stable walking on uneven terrains \cite{caron2019capturability}.

{To further enhance the walking robustness, Englsberger et al. proposed a measurement-based tracking controller to integrate vertical CoM motion with body rotation \cite{englsberger2012integration}. Utilizing the general Nonlinear Inverted Pendulum plus Flywheel Model (NIPFM) \cite{kajita2007human},} Lack et al. \cite{lack2015integrating} and Shafiee-Ashtiani et al. \cite{shafiee2017robust} studied the upper-body rotation and vertical height variation using the LMPC framework. However, they merely followed the pre-defined height trajectory. Based on the work in \cite{caron2019capturability}, Guan et al. proposed the Virtual-mass-ellipsoid
Inverted Pendulum (VIP) model \cite{guan2019push,guan2019virtual}. Then, they enhanced the intrinsically stable MPC for humanoid gait generation \cite{scianca2016intrinsically} and realized the disturbance rejection during stairs climbing. Nevertheless, to the best of our knowledge, there is still a lack of research that integrates step location adjustment, angular momentum adaptation and CoM height variation simultaneously. 

Differing from the above work, we propose an NMPC-based walking pattern generator for robust locomotion. By considering the dynamics effects caused by the change of roll and pitch angular momentum and the CoM height variation, the proposed approach can generate stable walking patterns by merely using walking parameter references. Based on state feedback, this optimizer is capable of modulating CoM height trajectory and body inclination angle in real-time instead of strictly tracking the pre-defined values.  

{The main contributions can be concluded as follows. Firstly, the manipulation of the ZMP trajectory (ankle strategy), the adjustment of step location (stepping strategy), the change of angular momentum (hip strategy), and the variation of CoM height (height variation strategy) are integrated into one single NMPC. As a result, a versatile framework for walking pattern generator is proposed, which in turn dramatically enhances the robustness in compensation for severe external disturbances. Secondly, through employing body rotation and CoM vertical movement, higher adaptability in real-world scenarios is also achieved. For example, the robot can pass through the limited space or climb higher stairs by modifying the CoM height and body rotation state in real-time. Thirdly, based on the NIPFM dynamics, the nonlinear constrained optimization can be transformed as a QCQP problem, which is solved fast via the SQP algorithm.}

The rest of this paper is organized as follows. In Section \ref{perliminary}, we briefly review the CoM dynamics and the procedure of SQP for NMPC solution. In Section \ref{formulation}, the QCQP problem is formulated. In \ref{NIPFM simulation} and \ref{whole-body-simu}, the inverted pendulum simulations and whole-body dynamic simulations are conducted. Finally, in Section \ref{conclusion}, we draw conclusions.

\section{Preliminaries} \label{perliminary}

This section introduces the fundamental knowledge about the NIPFM dynamics, the general MPC framework, the QCQP theory, and the SQP algorithm.

\subsection{NIPFM Dynamics} \label{CoM dynamics}
The LIPM \cite{kajita1991study}, which is a linear approximation of humanoid walking dynamics, assumes 1) the robot has a lumped mass body; 2) legs are mass-less and telescopic; 3) {CoM moves in a constant plane}. These assumptions, which over-constrain the robot's motion capabilities, limit the robot's performance undergoing external perturbations. In this work, in order to enhance the bipedal mobility, we propose to use the NIPFM to utilize the whole-body ability, especially the upper-body inclination and vertical CoM height variation, for robust locomotion. 

\begin{figure*} 
  \centering
  \includegraphics[width=120mm]{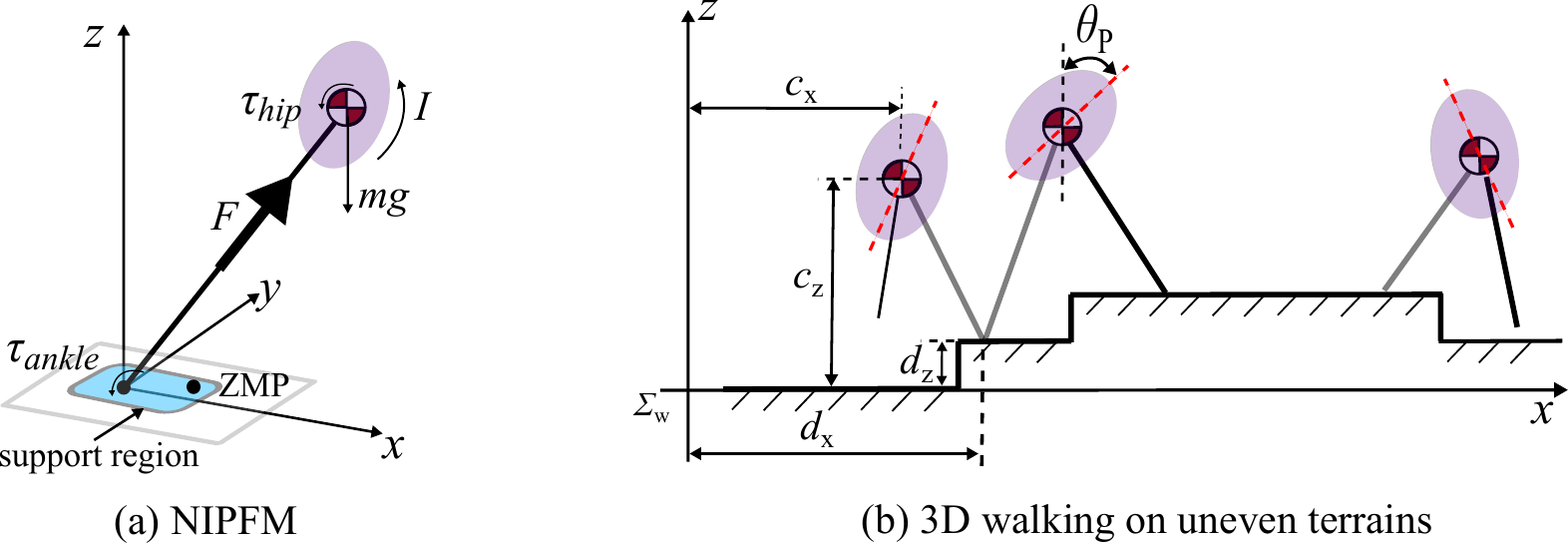}
  \caption{NIPFM for bipedal walking. {(a) The ankle torque helps to manipulate the ZMP and the flywheel rotation models the effect of the change of angular momentum. Note that the vertical height is changeable.}  (b) The definition of step parameters when walking on uneven terrains.}
  \label{fig:robot_fly_wheel1}
\end{figure*}

As demonstrated in Fig. \ref{fig:robot_fly_wheel1}, the NIPFM (referred as VIP in \cite{guan2019push,guan2019virtual}),  assumes 1) the flywheel has rotational inertia; 2) legs are mass-less and telescopic; 3) the CoM is located at the hip joint; 4) the CoM moves arbitrarily as long as physical limits are satisfied. {Thus, it can be used to model upper-body rotation and vertical body motion. Using this model, the ZMP, which should be restricted inside the robot's support polygon,} is computed by \cite{kudoh2002dynamic,kajita2007human,wieber2016modeling}
\begin{equation} \label{eq:IPFMmodel1}
\begin{aligned}
{p_{\text{x}}} &= c_{\text{x}} - \frac {c_{\text{z}}-d_{\text{z}}} {g+\ddot{c}_{\text{z}}} \ddot{c}_{\text{x}} - \frac{\dot{L}_{\text{y}}}{m(g+\ddot{c}_{\text{z}})},\dot{L}_{\text{y}} = I_{\text{y}} \ddot{\theta}_\text{p},\\
{p_{\text{y}}} &= c_{\text{y}} - \frac {c_{\text{z}}-d_{\text{z}}} {g+\ddot{c}_{\text{z}}} \ddot{c}_{\text{y}} + \frac{\dot{L}_{\text{x}}}{m(g+\ddot{c}_{\text{z}})},\dot{L}_{\text{x}} = I_{\text{x}} \ddot{\theta}_\text{r},\\
{{p_{\text{z}}}} &{= d_{\text{z}}},
\end{aligned}
\end{equation}
\normalsize
where $[p_{\text{x}}, p_{\text{y}}, p_{\text{z}}]^T$, $[c_{\text{x}}, c_{\text{y}}, c_{\text{z}}]^T$ and $[d_{\text{x}}, d_{\text{y}}, d_{\text{z}}]^T$ denote the 3D ZMP position, CoM position and supporting foot location, respectively. $[L_{\text{x}}, L_{\text{y}}]^T$, $[I_{\text{x}}, I_{\text{y}}]^T$ and $[{\theta}_\text{r}, {\theta}_\text{p}]^T$ separately denote the {(roll and pitch) angular momentum}, moment of inertia, and flywheel rotation angle about $x$- and $y$- axis ($x$- axis pointing to the forward direction in the sagittal plane and $y$- axis pointing to the left direction in the coronal plane). $m$ is the total mass, and $g$ is the gravitational acceleration.

\subsection{MPC State Equation} \label{framework}

{To build a MPC, the state equation should be first established. In this work, inspired by the work in \cite{kajita2003biped,wieber2006trajectory,diedam2008online}, we choose to use the Euler-integration equation}. Specifically, assuming the constant jerk of CoM movement and body rotation over the time interval $\Delta t$, we can compute the next state at time $t_{k+1}$,
\begin{equation} \label{eq:MPC_STATE}
\begin{aligned}
\hat{x}_{(k+1)} = \mathbf{A}\hat{x}_{(k)} + \mathbf{B}\dddot{x}_{(k)},
\end{aligned}
\end{equation}
\normalsize
where $\hat{x}_{(k)} =[x_{(k)}, \dot{x}_{(k)},\ddot{x}_{(k)}]^T$ ($x \in \{c_{\text{x}}, c_{\text{y}}, c_{\text{z}}, \theta_{\text{r}}, \theta_{\text{p}} \}$) is the current state, $\dddot{x}_{(k+1)}$ is the expected jerk,
\begin{equation} \nonumber
\mathbf{A}=
\left[ \begin{array}{ccc}
1 & \Delta t & \frac{1}{2}\Delta t^2\\
0 & 1 & \Delta t\\
0 & 0 & 1
\end{array} 
\right ],
\mathbf{B}=
\left[ \begin{array}{ccc}
\frac{1}{6}\Delta t^3\\
\frac{1}{2}\Delta t^2\\
\Delta t
\end{array}
\right ].
\end{equation}

Using (\ref{eq:MPC_STATE}), we can derive relationships between the jerk, its position, velocity and acceleration over the prediction horizon (of length $N_h$),
\begin{equation} \label{eq:MPC_state_re}
\begin{aligned}
{\mathbf{X}}_{(k)} &= \mathbf{P}_{\text{ps}}\hat{x}_{(k)} + \mathbf{P}_{\text{pu}}\dddot{\mathbf{X}}_{(k)}, \\
\dot{\mathbf{X}}_{(k)} &= \mathbf{P}_{\text{vs}}\hat{x}_{(k)} + \mathbf{P}_{\text{vu}}\dddot{\mathbf{X}}_{(k)},\\
\ddot{\mathbf{X}}_{(k)} &= \mathbf{P}_{\text{as}}\hat{x}_{(k)} + \mathbf{P}_{\text{au}}\dddot{\mathbf{X}}_{(k)},
\end{aligned}
\end{equation}
\normalsize
where
${\mathbf{X}}_{(k)}=[x_{(k+1)},...,x_{(k+N_h)}]^T$ ($\mathbf{X} \in \{\mathbf{C}_{\text{x}}, \mathbf{C}_{\text{y}}, \mathbf{C}_{\text{z}} \mathbf{\mathbf{\Theta}}_{\text{r}}, \mathbf{\mathbf{\Theta}}_{\text{p}} \}$) denotes the future state of CoM along $x$-, $y$- and $z$- axis and the upper-body inclination state about $x$- and $y$- axis over the prediction horizon, where, e.g. $\mathbf{C}_{\text{x}(k)}=[c_{\text{x}(k+1)},...,c_{\text{x}(k+N_h)}]^T$. $\dot{\mathbf{X}}_{(k)}$, $\ddot{\mathbf{X}}_{(k)}$, and $\dddot{\mathbf{X}}_{(k)}$ separately denote the velocity, acceleration and jerk of each motion channel, respectively. ${\mathbf{X}}_{(k)}=[x_{(k+1)},...,x_{(k+N_h)}]^T$, and separately. $\mathbf{P}_{\text{ps}}$, $\mathbf{P}_{\text{pu}}$, $\mathbf{P}_{\text{vs}}$, $\mathbf{P}_{\text{vu}}$, $\mathbf{P}_{\text{as}}$ and $\mathbf{P}_{\text{au}}$ are obtained by calculating (\ref{eq:MPC_STATE}) recursively \cite{diedam2008online}.


\subsection{SQP-based QCQP solution}
{Using the state equation expressed in \eqref{eq:MPC_STATE} as the predictive model, the MPC problem is built as a QCQP, when considering the feasible constraints, especially the ZMP stability constraint.} Specifically, the QCQP can be expressed as follows:
\begin{align}
\mathop{\min_{\bm{\mathcal{X}}}} \quad f(\bm{\mathcal{X}}) & = \bm{\mathcal{X}}^T \mathbf{G}\bm{\mathcal{X}} + \mathbf{g}^T\bm{\mathcal{X}}, \nonumber \label{eq:qcqp} \\
\text{s.t.} \quad h_j(\bm{\mathcal{X}}) & \leq 0, \\
h_j(\bm{\mathcal{X}})  & = \bm{\mathcal{X}}^T \mathbf{V}_j \bm{\mathcal{X}} + \mathbf{v}_{j}^T \bm{\mathcal{X}}+ \sigma_j, j \in \{1,...,N_c\}, \nonumber
\end{align}
where $\bm{\mathcal{X}} \in \Re^{N_t}$ is the state vector, $N_c$ and $N_t$ are the number of constraints and state variables, respectively. $\mathbf{G}$, $\mathbf{V}_j \in \Re^{N_t \times N_t}$, $\mathbf{g}$, $\mathbf{v}_j \in \Re^{N_t}$, and $\sigma_j \in \Re$ are the parameters that specify the objective function and constraints. 

The above nonlinear optimization problem can be solved by the SQP algorithm \cite{nocedal2006sequential}. By using the SQP, the solution of \eqref{eq:qcqp} is transformed as solving following problem,
\begin{equation} \label{eq:sqp}
\begin{aligned}
\mathop{\min_{\Delta_{\bm{\mathcal{X}}}}} \quad \frac{1}{2} \Delta_{\bm{\mathcal{X}}}^T \nabla_{\bm{\mathcal{X}}}^2(f(\bm{\mathcal{X}})) \Delta_{\bm{\mathcal{X}}} &+ (\nabla_{\bm{\mathcal{X}}} (f(\bm{\mathcal{X}})))^T \Delta_{\bm{\mathcal{X}}}, \\
\text{s.t.} \quad \quad  (\nabla_{\bm{\mathcal{X}}} (h_j(\bm{\mathcal{X}})))^T \Delta_{\bm{\mathcal{X}}} &+h_j(\bm{\mathcal{X}}) \leq 0,\\
\end{aligned}
\end{equation}
\normalsize
where {$\nabla$ and $\nabla^2$ separately denote the first-order and the second-order differential operator, respectively}.

By using \eqref{eq:qcqp}, we can easily obtain
\begin{equation} \label{differential_eq}
\begin{aligned}
\nabla_{\bm{\mathcal{X}}}^2 (f (\bm{\mathcal{X}})) &= 2\mathbf{G}, &&\nabla_{\bm{\mathcal{X}}} (f (\bm{\mathcal{X}}))= 2\mathbf{G} \bm{\mathcal{X}} + \mathbf{g}, \\
\nabla_{\bm{\mathcal{X}}}^2 (h_j(\bm{\mathcal{X}})) \!\! & = 2\mathbf{V}_j, && \nabla_{\bm{\mathcal{X}}} (h_j(\bm{\mathcal{X}}))= 2\mathbf{V}_j \bm{\mathcal{X}} + \mathbf{v}_j. \\
\end{aligned}
\end{equation}
\normalsize 

{As a result, the QCQP can be transformed as the QP, which can be solved by time-efficient algorithms such as the active-set method. The solution of the QP ($\Delta_{\bm{\mathcal{X}}} \in \Re^{N_t})$ is then used to update the state variable via $\bm{\mathcal{X}} = \bm{\mathcal{X}} + \Delta_{\bm{\mathcal{X}}}$. Once completed, the solving process of (\ref{eq:sqp}) is repeated with the new $\bm{\mathcal{X}}$ for $N_s$ times until meeting the convergence condition, which will be discussed in detail in Section \ref{compute-effi}.}

\section{Problem Formulation} \label{formulation}
{Considering the NIPFM dynamics and feasibility limitations, a NMPC problem is established and solved to generate versatile walking patterns by exploring the step location adjustment, angular momentum adaptation and vertical height variation}. This section discusses the objective function and the feasibility constraints in detail.   
\subsection{Objective Function} \label{objective}
After defining the reference step location and walking cycle, the work aims to minimize the tracking errors of CoM positions, upper-body inclination angles, and step locations. Also, we minimize the velocities and jerks of the CoM movement and body inclination. 
Thus, at the $k^{\text{th}}$ sampling time, we have the objective function as follows:
\begin{equation} \label{eq:obj_Model}
\begin{aligned}
f\!&= \!
\sum_{\mathbf{X}}
\Big\{ \!\frac{\alpha_{_{\mathbf{X}}}}{2}\! \parallel \!\dot{\mathbf{X}}_{(k)}\!\! \parallel^2 \!+\! \frac{\beta_{_{\mathbf{X}}}}{2} \!\parallel \!{\mathbf{X}}_{(k)} \!-\! {\mathbf{X}}_{(k)}^{\text{ref}} \! \parallel^2
 \!+\! \frac{\gamma_{_{\mathbf{X}}}}{2} \parallel \! \dddot{\mathbf{X}}_{(k)} \!\parallel^2 \!\!\Big\} \\
&+\!\sum_{\mathbf{U}} \!\frac{\delta_{_{\mathbf{U}}}}{2} \! \parallel \!\! \mathbf{U}_{(k)} \!-\! \mathbf{U}_{(k)}^{\text{ref}} \! \parallel^2.
\end{aligned}
\end{equation}
\normalsize
where $\alpha_{_{\mathbf{X}}}$, $\beta_{_{\mathbf{X}}}$, $\gamma_{_{\mathbf{X}}}$ and $\delta_{_{\mathbf{U}}}$ separately denote the velocity, position tracking, jerk, support position tracking  penalties.  ${\mathbf{X}}_{(k)}^{\text{ref}}=[x_{(k+1)}^{\text{ref}},...,x_{(k+N_h)}^{\text{ref}}]^T$ are the reference states of CoM position and upper-body inclination angles over the prediction horizon. {$\mathbf{U}_{(k)}=[u_{(k, 1)},...,u_{(k, N_f)}]^T$ and $\mathbf{U}_{(k)}^{\text{ref}} =[u_{(k, 1)}^{\text{ref}},...,u_{(k, N_f)}^{\text{ref}}]^T$ are the actual and reference future step locations over the prediction horizon ($\mathbf{U}_{(k)} \in \{ {\mathbf{D}_{\text{x}}}, {\mathbf{D}_{\text{y}}},{\mathbf{D}_{\text{z}}}\}$). $[\mathbf{D}_{\text{x}(k)}, \mathbf{D}_{\text{y}(k)}, \mathbf{D}_{\text{z}(k)}]^T$ 
are the actual horizontal step locations over the prediction horizon, where, e.g. ${\mathbf{D}_{\text{x}(k)} =[d_{\text{x}(k, 1)},...,d_{\text{x}(k, N_f)}]^T} $}
\footnote{In this paper, we use $[d_{\text{x}(k, 1)},...,d_{\text{x}(k, N_f)}]^T$ to denote the future step locations of different walking cycles falling on the prediction horizon, and $[d_{\text{x}(k+1)},...,d_{\text{x}(k+N_h)}]^T$ to denote step locations at different sampling
times over the prediction horizon.}. $N_f$ is the number of future step locations over the prediction horizon. Note that in this work, the actual step height ($\mathbf{D}_{\text{z}(k)}$) is set to be the desired step height ($\mathbf{D}_{\text{z}(k)}^{\text{ref}}$), which is determined by the surface height configuration.

{Under this objective function, the parameters for (\ref{eq:qcqp}) can be calculated by using the predictive model expressed in \eqref{eq:MPC_STATE} and \eqref{eq:MPC_state_re}}. More details can be found in Appendix A. 

{During each walking cycle, the reference CoM positions along $x$- and $y$- axis are set to be the center of the reference step locations, as explained in Appendix B. By default, the reference roll angles and pitch angles are set to be zero during the whole walking process so that the robot keeps upright. Particularly, the reference CoM height is the sum of the default inverted pendulum height (denoted by $h_\text{z}^{	\text{ref}}$) and the reference step height. That is,}
\begin{equation} \label{eq:footst_reference}
\left\{
\begin{aligned}
\theta_{\text{r}(k+i)}^{\text{ref}} &=\!\theta_{\text{p}(k+i)}^{\text{ref}}= 0,  && i \in \{1,...,N_h \},\\ 
c_{\text{z}(k+i)}^{\text{ref}} &= h_\text{z}^{\text{ref}} + d_{\text{z}(k+i)}^{\text{ref}}, &&i \in \{1,...,N_h \},\\
\end{aligned}
\right .\,
\end{equation}
\normalsize
where $[\theta_{\text{r}(k+i)}^{\text{ref}},...,\theta_{\text{r}(k+N_h)}^{\text{ref}}]^T$ and $[\theta_{\text{p}(k+i)}^{\text{ref}},...,\theta_{\text{p}(k+N_h)}^{\text{ref}}]^T$ denote the reference roll angles and pitch angles over the prediction horizon,  {$h_\text{z}^{\text{ref}}$ is the default inverted pendulum height determined by physical structure}, $[c_{\text{z}(k+i)}^{\text{ref}},...,c_{\text{z}(k+N_h)}^{\text{ref}}]^T$ denote the reference CoM height, $[d_{\text{z}(k+i)}^{\text{ref}},...,d_{\text{z}(k+N_h)}^{\text{ref}}]^T$ denote the reference step height which is determined by the surface height configuration.

\subsection{Constraints} \label{constraints}
In order to guarantee the feasibility, this work takes into account the constraints on the ZMP movement, step location variation, CoM vertical motion, upper-body inclination, and hip torque output. 

\subsubsection{Constraints on ZMP trajectory} The ZMP should be restricted inside the support polygon in order to maintain the balance. Although the support polygon expands when switching from single supporting phase to double supporting phase, we only consider the single supporting phase during the whole walking process. Since those ZMP movement constraints applied on single supporting phase are more restrictive at this stage and the sampling time can be large enough in the MPC framework\cite{diedam2008online}, this simplification is reasonable. Therefore, at the $k^{\text{th}}$ sampling time, taking the sagittal movement for an example, the following constraint needs to be satisfied:
\begin{equation} \label{eq:zmp_constrain}
\begin{aligned}
     p_\text{x}^{\min} \leq p_{\text{x}(k+i)} -d_{\text{x}(k+i)} \leq p_\text{x}^{\max}, i \in \{1,...,N_h \},
\end{aligned}
\end{equation}
\normalsize
where $[p_{\text{x}(k+1)},...,p_{\text{x}(k+N_h)}]^T$ denote the generated ZMP position over the prediction horizon along the $x$- axis,  $[d_{\text{x}(k+1)},...,d_{\text{x}(k+N_h)}]^T$ denote the generated step location over the prediction horizon along the $x$- axis, $p_\text{x}^{\min}$ and $p_\text{x}^{\max}$ are the lower and upper ZMP boundary along $x$- axis, which are determined by the size of supporting foot.  

The ZMP movement in the coronal plane should also satisfy the similar constraints. When using CoM dynamics expressed in (\ref{eq:IPFMmodel1}), the ZMP constraints are nonlinear inequalities. 
As demonstrated in Appendix B,  they can be formulated as quadratic forms.

\subsubsection{Constraints on step location variation} \label{foot constraints}
The objective function takes the horizontal step locations as variables that can change arbitrarily. However, such modification should satisfy physical limitations, such as the maximal leg stretch length, motor actuation capability, self-collision avoidance etc. In this paper, the following limitations are considered.

Firstly, due to limitations arisen from the mechanical structure  and actuation capability,  the range of step parameters, including step length and step width, should be constrained. At present, these limitations are simplified to be following linear inequalities (taking the forward movement for an example):
\begin{equation} \label{eq:footstep_constrain}
\left\{
\begin{aligned}
d_\text{x}^{\min} &\!\leq d_{\text{x}(k,i)} \!-\!\hat{d}_{\text{x}(k)} \!\leq d_\text{x}^{\max},  && i = 1,\\ 
d_\text{x}^{\min} &\!\leq d_{\text{x}(k,i)} \!-\!d_{\text{x}(k,i-1)} \!\leq d_\text{x}^{\max}, &&i \in \{2,...,N_f \},\\
\end{aligned}
\right .\,
\end{equation}
\normalsize
where $\hat{d}_{\text{x}(k)}$ denotes the current supporting foot position along the $x$- axis, $d_\text{x}^{\min}$ and $d_\text{x}^{\max}$ are lower and upper boundaries of step length. 



Secondly, since the future step location falls into the same walking cycle is re-computed in each loop, it may change in real-time. However, the position of swing foot, which is determined by the future step location, cannot change rapidly due to the joint velocity limitation. In this work, for the simplification, rather than constraining the swing foot trajectory, we limit the change of future step locations corresponding to the same walking cycle generated by different control loops. Since this constraint is imposed at each time interval, limiting the next one future step location is enough. Thus, we have (taking the forward motion for an example):
\begin{equation} \label{eq:footstep_v_constrain}          
\begin{aligned}
\dot{d}_{\text{x}}^{\min} \Delta t \leq d_{\text{x}(k, 1)} - d_{\text{x}(k-1, 1)} \leq \dot{d}_{\text{x}}^{\max} \Delta t,
\end{aligned}
\end{equation}
\normalsize
where $d_{\text{x}(k, 1)}$ and $d_{\text{x}(k-1, 1)}$ are the next one step positions corresponding to the same period computed by current and last optimization loop, $\dot{d}_{\text{x}}^{\min}$ and $\dot{d}_{\text{x}}^{\max}$ are the lower and upper velocity boundaries along the $x$- axis. 

Finally, as mentioned above, the actual step height is merely determined by the surface height configuration. Thus, we have the equality constraint on step height:
\begin{equation} \label{eq:foot height_constrain}
\begin{aligned}
     d_{\text{z}(k,i)}=d_{\text{z}(k,i)}^{\text{ref}}, i \!\in \!\{1,...,N_f \},
\end{aligned}
\end{equation}
\normalsize
where $d_{\text{z}(k,i)}^{\text{ref}}$ is the reference step height determined by the surface height.

\subsubsection{Constraints on CoM motion} \label{CoM constraints}
{According to \eqref{eq:IPFMmodel1}, the vertical height variation can be utilized to manipulate the ZMP movement}, but the height variation should be constrained so as to avoid infeasible trajectories which break the kinematic constraints. The limitation of the CoM vertical motion leads to the following constraints:
\begin{equation} \label{eq:height_constrain}
\begin{aligned}
     h^{\min} \!\leq \!c_{\text{z}(k+i)} \!-\!d_{\text{z}(k+i)} \!-\!h_z^{\text{ref}} \!\leq h^{\max}, i \!\in \!\{1,...,N_h \},
\end{aligned}
\end{equation}
\normalsize
where $h^{\min}$ and $h^{\max}$ are the lower and upper boundaries of the vertical height variance, $[d_{\text{z}(k+1)},...,d_{\text{z}(k+N_h)}]^T$ denote the generated vertical height of the supporting foot over the prediction horizon.

Additionally, since the ground only generates unilateral reactive forces, the vertical acceleration should be limited to avoid free fall. That is:
\begin{equation} \label{eq:height_vertical_constrain}
\begin{aligned}
 \ddot{c}_{\text{z}(k+i)} \geq -g, i \in \{1,...,N_h \}.
\end{aligned}
\end{equation}
\normalsize

\subsubsection{Constraints on body inclination}
The trunk rotation is limited by articulation constraints. When using NIPFM, it can be constrained by using following bounds (taking the roll angle for an example):
\begin{equation} \label{eq:angle_constrain}
\begin{aligned}
{\theta}_{\text{r}}^{\min} \leq {\theta}_{\text{r}(k+i)} \leq {\theta}_{\text{r}}^{\max}, i \in \{1,...,N_h \},
\end{aligned}
\end{equation}
\normalsize
where $\theta_{\text{r}}^{\min}$ and $\theta_{\text{r}}^{\max}$ are the lower and upper boundaries of roll angle.

\subsubsection{Constraints on hip joint torque} The hip joint torque should also be limited. Taking the roll direction for an example:
\begin{equation} \label{eq:torque_constrain}
\begin{aligned}
\tau_{\text{r}}^{\min} \leq I_{\text{x}} \ddot{\theta}_{\text{r}(k+i)} \leq \tau_{\text{r}}^{\max}, i \in \{1,...,N_h \}, \\
\end{aligned}
\end{equation}
\normalsize
where $\tau_{\text{r}}^{\min}$ and $\tau_{\text{r}}^{\max}$ are the lower and upper limits of roll torque.

\section{Inverted Pendulum Simulations} \label{NIPFM simulation}
Here, we validate the effectiveness of the proposed algorithm, where the bipedal walking in multiple scenarios, including 3D walking across uneven terrains and push recovery from external disturbances, is demonstrated. Particularly, this section focuses on the inverted pendulum simulations by using the physical characteristics of the COMAN humanoid robot\cite{tsagarakis2013compliant}. The robot body has 25 DoFs. The total weight is 31 kg and the total height is 0.95 m. For the NIPFM simulation, the fixed time interval $\Delta t$ is 0.005 s and the predictive length $N_h$ is 31. That is, the predictive window is 1.55 s long. For the gait generation, the constant walking cycle ($T$) is 0.8 s. As a result, the information of the two future steps is utilized in each optimization loop. Other essential parameters for the inverted pendulum simulations are listed in Appendix C. 

\subsection{3D walking across uneven terrains} \label{natural}
\begin{figure}
  \centering
  \includegraphics[width=\columnwidth-1mm]{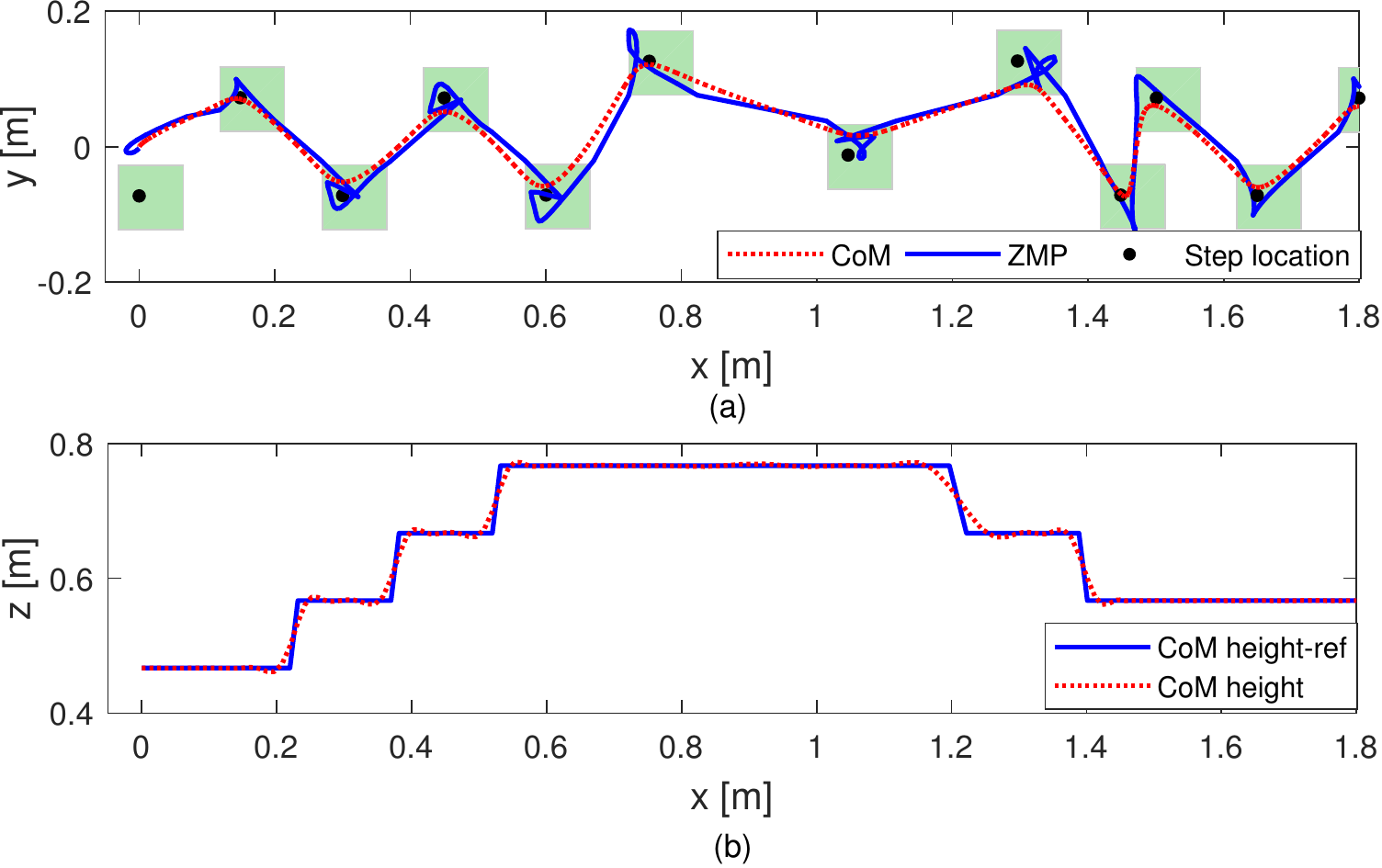}
  \caption{{Bipedal gait generation for 3D walking across uneven terrains. (a) Generated horizontal CoM trajectory, ZMP trajectory and step locations. The green blocks represent the supporting foot area. (b) Generated sagittal CoM trajectory. Defining the constant height reference during each walking cycle, time-varying height trajectory is automatically generated.}}
  \label{fig:3D walking-horizontal}
\end{figure}

\begin{figure}
  \centering
  \includegraphics[width=\columnwidth-1mm]{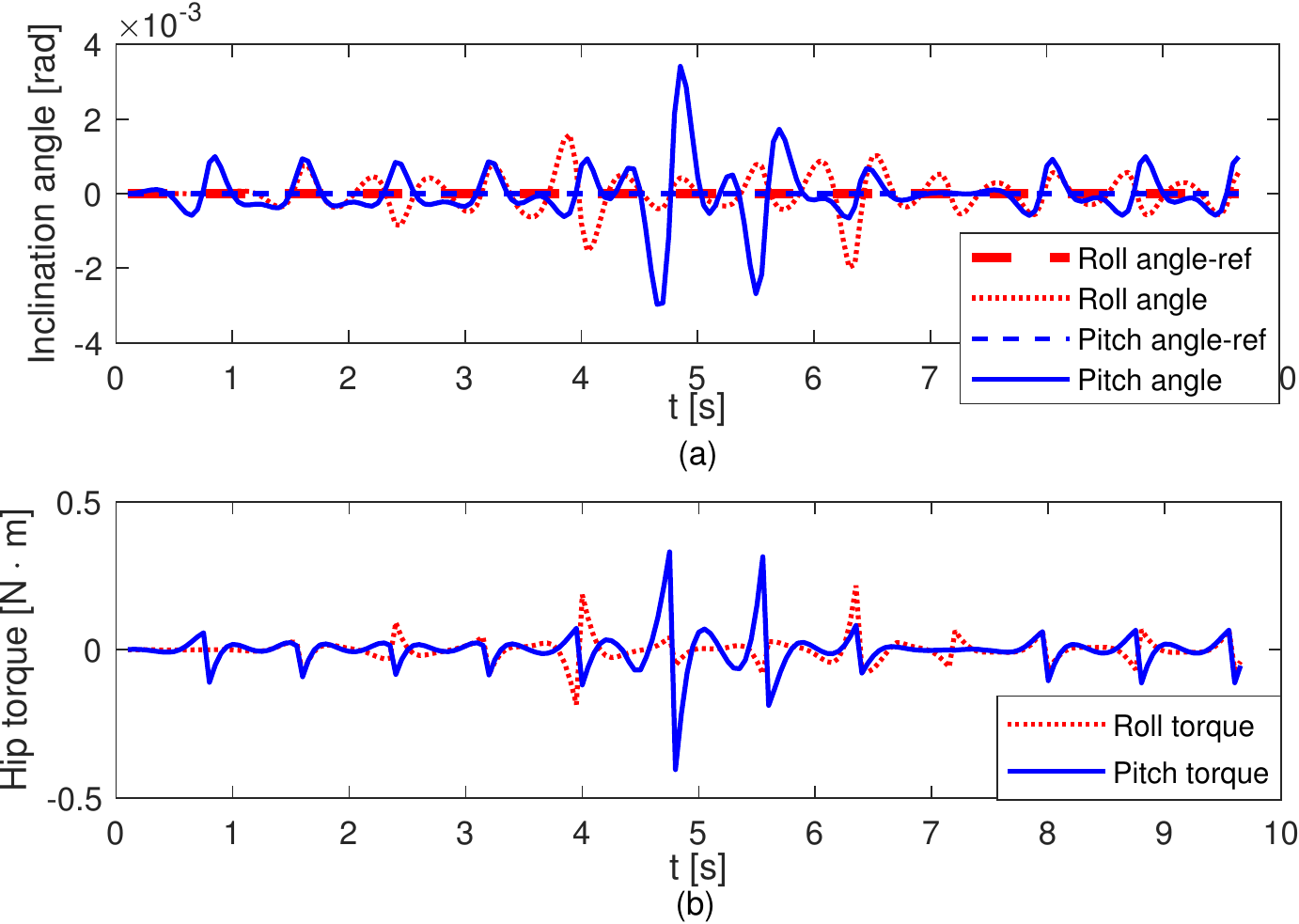}
  \caption{{Generated body inclination angles (a) and hip torques (b) for 3D walking across uneven terrains. Since the CoM height variation is integrated, the generated body rotation angles and hip torques are very low.} }
  \label{fig:3D walking-angle}
\end{figure}

In this section, the performance of the proposed NMPC framework is tested in a 3D walking scenario where the robot {is expected to walk across uneven terrains, including upstairs and downstairs. To further validate the walking adaptability, the step parameters (including step length and step width) variation is also taken into consideration. The reference step parameters are listed in Table \ref{table:step param for 3d walking}. The results are illustrated in Fig. \ref{fig:3D walking-horizontal} and Fig. \ref{fig:3D walking-angle}.} 

\begin{table}
\caption{Reference step parameters for 3D walking on uneven terrains.}
\centering
\setlength{\tabcolsep}{1mm}
{\begin{tabular}{lcccccccc} \toprule
 \diagbox{Param.}{Cycle} & 1 & 2-4 & 5 & 6 & 7 &8&9& 10- \\ \midrule
 Step length[m] & 0.15 &0.15 & 0.15 & 0.3 & 0.25 & 0.15 & 0.05& 0.15\\
 Step width[m] & 0.145 & 0.145 & 0.2 & 0.14 & 0.14 &0.2& 0.145& 0.145 \\
 Step height[m]  & 0 & 0.1 & 0 & 0 & -0.1 & -0.1& 0& 0 \\
 \bottomrule
\end{tabular}}
\label{table:step param for 3d walking}
\end{table}

As illustrated in Fig. \ref{fig:3D walking-horizontal} (a), through restricting the ZMP trajectory within the allowable region formed by the supporting feet, the proposed method is capable of generating  3D feasible gait which satisfies the stability constraints. {Further observations demonstrate that the NMPC algorithm can manipulate the ZMP movement to maintain the balance, even when the step length and step length vary dramatically from the $5^{\text{th}}$ cycle to the $9^{\text{th}}$ cycle. Namely, the ankle strategy \cite{aftab2012ankle} is utilized by the proposed NMPC framework.} 

As can be seen in Fig. \ref{fig:3D walking-horizontal} (b), vertical CoM motion is also employed when walking across the uneven terrains. In this case, it should be noted that we only take the reference surface height as the input and assume the constant pendulum height ($h_\text{z}^{\text{ref}}$) during the whole process. Nevertheless, by using the proposed NMPC strategy, the time-varying height trajectory is automatically generated. {In addition, as can be seen in Fig. \ref{fig:3D walking-angle} (a), the robot slightly rotates the upper-body for maintaining stability when stepping upstairs and downstairs, meaning that the angular momentum adaption is integrated. Due to the integration of the vertical body motion, the upper-body inclination is suppressed.} As a result, as demonstrated in Fig. \ref{fig:3D walking-angle} (b), the hip torque variation is also suppressed into a very low level.

\subsection{Balance recovery from external disturbances} \label{robust}
In this section, we analyse the push recovery capability of the proposed NMPC scheme. When walking in free space, human beings tend to take reactive steps to overcome the large disturbances. In this paper, {considering the upper-body rotation and height variation, three strategy combinations}, i.e., step location adjustment (strategy 1), step location adjustment plus upper-body rotation (strategy 2), and step location adjustment, upper-body rotation plus vertical height variation (strategy 3) are considered. In addition, when the external disturbances are smaller or when walking in certain circumstances where the step location adjustment is not allowed, the robot can only make use of the angular momentum adaptation and height variation\cite{guan2019push,horak1986central}. Thus, the upper-body rotation plus vertical height variation (strategy 4) is also discussed here.

{Using the objective function \eqref{eq:obj_Model}, above four strategy combinations can be activated flexibly by tuning the weight coefficients.} However, in this work, in order to compare the robustness of different strategy combinations in a quantitative manner, we apply equality constraints (hard constraints) on the generated step locations, CoM height trajectory, and body inclination angles to activate different strategy combinations. 

{In particular, the step location adjustment would not be enabled if following equality constraints are applied (taking the forward movement for an example):}
\begin{equation} \label{eq:foot_step_location_constrain}
\begin{aligned}
     d_{\text{x}(k,i)}=d_{\text{x}(k,i)}^{\text{ref}}, i \!\in \!\{1,...,N_f \},\\
\end{aligned}
\end{equation}
\normalsize
where $d_{\text{x}(k,i)}^{\text{ref}}$ is the forward step location reference.

The upper-body rotation would not be enabled if following equality constraints are applied (taking the roll rotation for an example):
\begin{equation} \label{eq:angle_eq_constrain}
\begin{aligned}
 {\theta}_{\text{r}(k+i)} = {\theta}_{\text{r}(k+i)}^{\text{ref}}, i \in \{1,...,N_h \}.
\end{aligned}
\end{equation}
\normalsize

The vertical height variation would not be enabled if following equality constraints are applied:
\begin{equation} \label{eq:comz_constrain}
\begin{aligned}
 c_{\text{z}(k+i)} = c_{\text{z}(k+i)}^{\text{ref}}, i \in \{1,...,N_h \}.
\end{aligned}
\end{equation}
\normalsize

Note that, to improve the compute efficiency in the real application, we can apply the above equality constraints \eqref{eq:angle_eq_constrain} and \eqref{eq:comz_constrain} merely on the next one sampling time.

\subsubsection{{Balance recovery when stepping in place}} \label{ipfm_walk_in_place}
\begin{figure*}
  \centering
  \includegraphics[width=130mm]{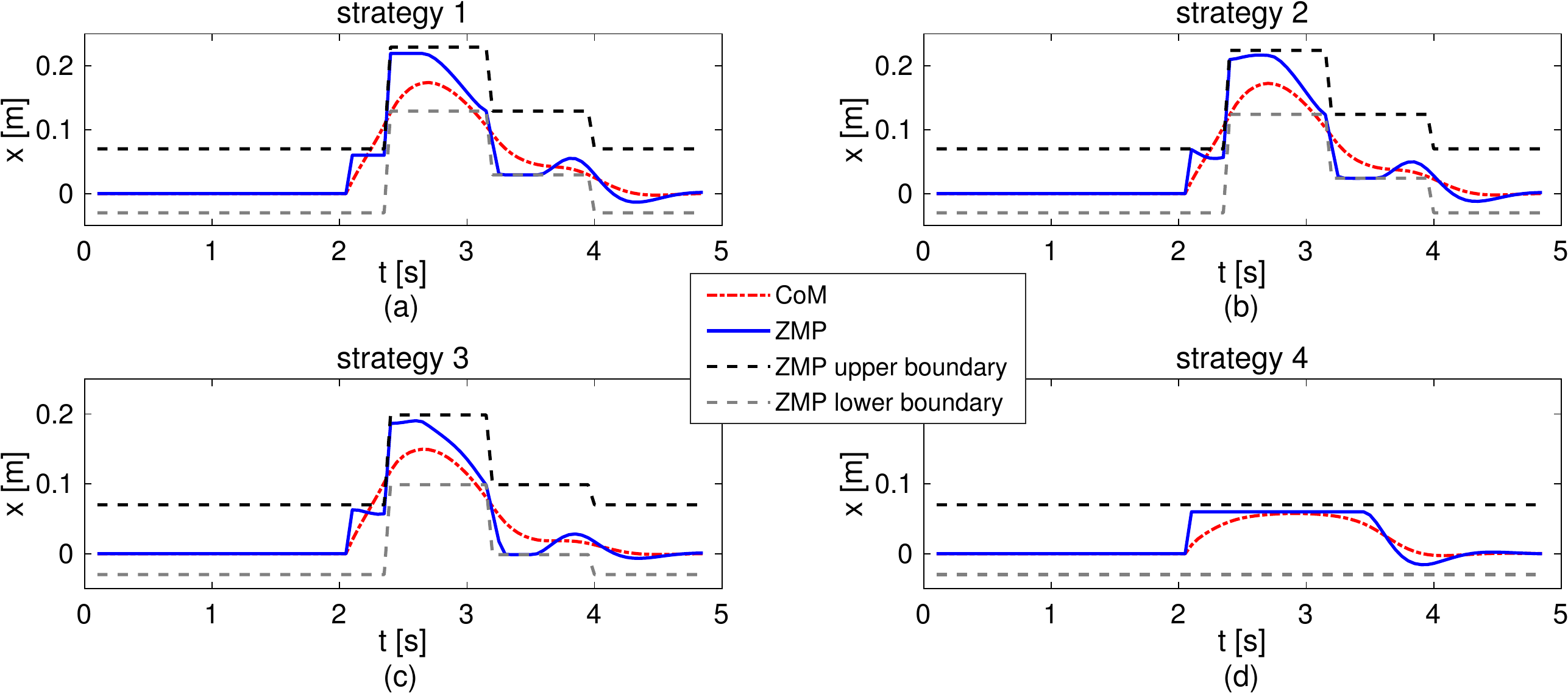}
  \caption{{Forward CoM trajectories, ZMP trajectories and ZMP boundaries generated by four different strategies. The ZMP boundaries are determined by the adjusted step locations.}}
  \label{fig:robuts1-stepping-position}
\end{figure*}

\begin{figure*}
  \centering
  \includegraphics[width=130mm]{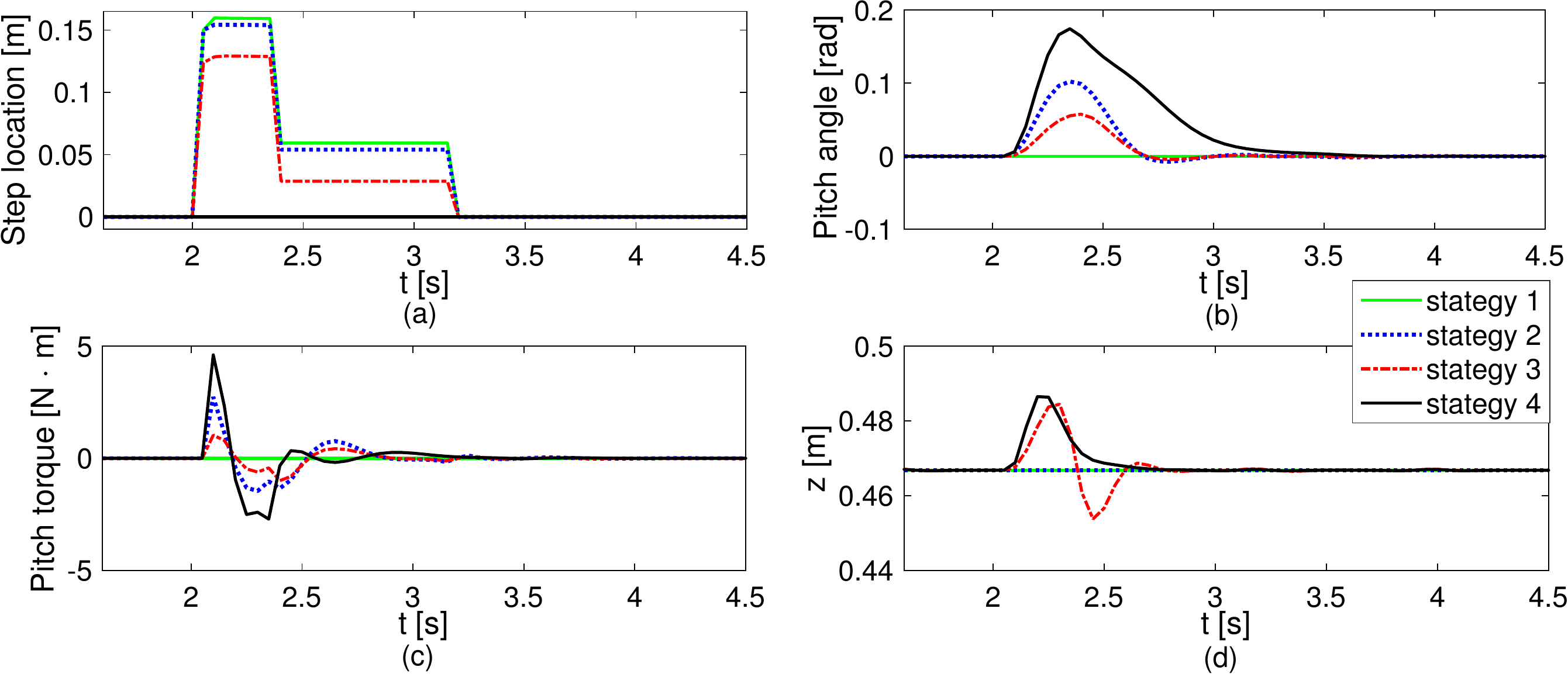}
  \caption{{Forward step location sequences (a), pitch angles (b), pitch torques (c) and CoM height trajectories (d) generated by four different strategies.} }
  \label{fig:robuts1-angle-position}
\end{figure*}

{In this section, the robot is expected to recover from external pushes when stepping in place.} Briefly, only the horizontal external force along the forward direction was applied at the pelvis at 2 s, lasting 0.1 s. Under the external force push, the walking patterns generated by four different strategies are demonstrated in Fig. \ref{fig:robuts1-stepping-position} and Fig. \ref{fig:robuts1-angle-position}. Note that 125 N forward force is applied when using strategy 1, 2 and 3, whereas only 80 N forward force is applied when using strategy 4.

As illustrated in Fig. \ref{fig:robuts1-stepping-position}, when using strategy 1, 2, and 3, the ZMP support region is extended by making reactive steps for disturbance rejection. {Taking the strategy 1 as an example,  the robot timely adjusts the step locations when the external force is imposed, as can be seen in Fig. \ref{fig:robuts1-angle-position} (a). Since the external disturbance is large enough, the next step length is also adjusted. As a result, the robot adjusts the step locations of the $4^{\text{th}}$ and $5^{\text{th}}$ steps (lasting from 2.4 s to 4.0 s), as illustrated in Fig. \ref{fig:robuts1-stepping-position} (a). Furthermore, since the objective function \eqref{eq:obj_Model} aims to track the global step locations, the step location returned to be 0 m again after the $6^{\text{th}}$ cycle}. The similar phenomena can also be found when using strategy 2 and strategy 3.

Besides, when using strategy 2 where merely the equality constraint expressed in \eqref{eq:comz_constrain} is applied, the angular momentum is also utilized. As demonstrated in Fig. \ref{fig:robuts1-angle-position} (b), the maximal pitch angle increases to be about 0.1 rad. Correspondingly, as can be seen in Fig. \ref{fig:robuts1-angle-position} (c), the pitch torque is needed to rotate the upper-body. When using strategy 3, where no constraint in \eqref{eq:foot_step_location_constrain}, \eqref{eq:angle_eq_constrain} or \eqref{eq:comz_constrain} is applied, the vertical height variation is also utilized to reject the external disturbance, as can be seen in Fig. \ref{fig:robuts1-angle-position} (d). {Compared with strategy 2, the maximal pitch angle decreases to be about 0.058 rad. In addition, the step length variation when using strategy 2 or strategy 3 is also reduced when compared with strategy 1. As can be seen in Fig. \ref{fig:robuts1-angle-position} (a), the maximal step length of strategy 1 is 0.16 m whereas the step length of strategy 2 and strategy 3 are 0.15 m and 0.13 m, respectively. That is to say, when walking in the narrow space with limited support zone, strategy 2 and strategy 3 contributes to high robustness against external disturbances.}

{As a comparison, strategy 4, where merely the equality constraint expressed in \eqref{eq:foot_step_location_constrain} is applied, is also tested in this section. As can be seen in Fig. \ref{fig:robuts1-stepping-position} (d), when the external push is imposed, step location adjustment is not utilized. Instead, the ZMP movement is utilized to reject the force perturbation. {At the same time, upper-body rotation and height variation are both utilized to maintain the balance, as demonstrated in Fig \ref{fig:robuts1-angle-position} (b) and (d). In this sense, when walking in the scenario such as stair climbing where the step location adjustment is not allowed, the upper-body rotation and height variation can be employed by the proposed method for disturbance rejection, which is similar with the work in \cite{guan2019virtual}}.
However, it should be mentioned that although only 80 N forward push is imposed when using strategy 4, which is smaller than the 125 N imposed on other strategies, the needed pitch angle and height variation by the strategy 4 are larger than those of other three strategy combinations, (see Fig. \ref{fig:robuts1-angle-position} (b) and (d)), meaning that the stepping strategy plays a key role in rejecting large disturbances.}

It is also worth mentioning that the ZMP trajectory moves within the support region, no matter which strategy combination is used, as can be seen in Fig. \ref{fig:robuts1-stepping-position}. That is to say, the ankle strategy is also utilized by the proposed NMPC scheme for disturbance rejection.

\subsubsection{Balance recovery when walking forward} \label{ipfm_forward_walking}

In this case, the robot is expected to recover from external pushes when walking forward. The horizontal external forces are applied at the pelvis at 2 s, lasting 0.1 s. The default step length is 0.15 m, and the default step width is 0.145 m. Other parameters are the same as those used in the last section. Particularly, multi-directional external forces are applied to the robot.

\begin{figure*}
  \centering
  \includegraphics[width=140mm]{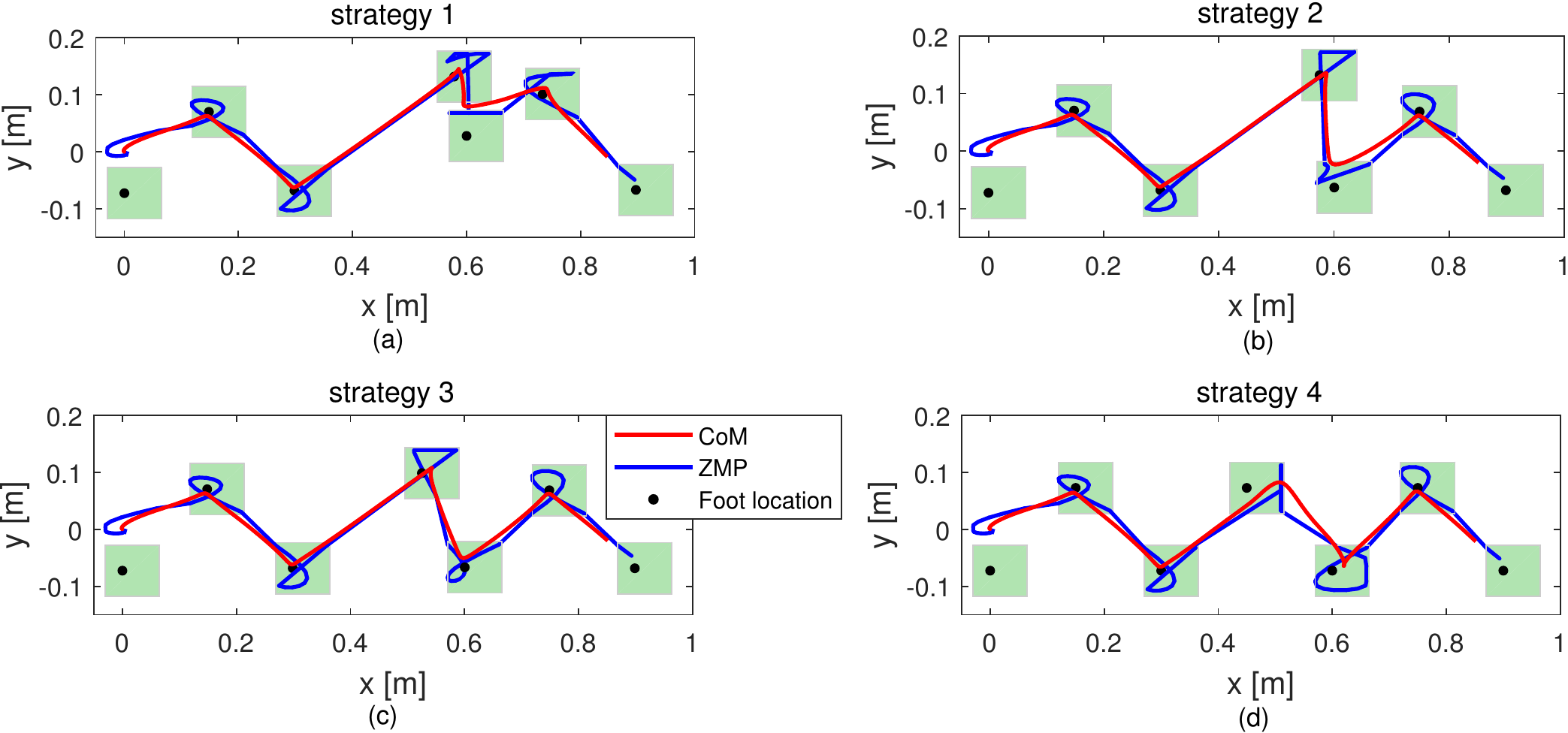}
  \caption{{Horizontal CoM trajectory, ZMP trajectory and step locations generated by different strategies for rejecting the same push force. The green blocks represent supporting foot plane.}}
  \label{fig:robuts1-position}
\end{figure*}

Under the same external push (forward 125 N, lateral 75 N in this case), the robust walking patterns generated by four different strategy combinations are demonstrated in Fig. \ref{fig:robuts1-position}. Similar to the result in the last section, when using strategy 1, which has been adopted in other work such as \cite{diedam2008online}, the robot also adjusts the step locations when the external push occurs. Numerical analysis shows that the step length is modified to be 0.28 m (note the maximal allowable step length is 0.3 m) and the step width is switched from 0.145 m to 0.19 m (note the maximal allowable step width is 0.2 m). To recover from the perturbation, the next step location is also adjusted (from (0.6 m,-0.0725 m) to (0.605 m,0.028 m)). After three steps, the robot fully recovers from the external push and tracks the reference step locations again. Further observation reveals that the ZMP trajectory goes along the border of the support region at the $4^{\text{th}}$ and $5^{\text{th}}$ steps. That is to say, the ankle strategy is also employed to keep stability.

\begin{figure}
  \centering
  \includegraphics[width=\columnwidth-1mm]{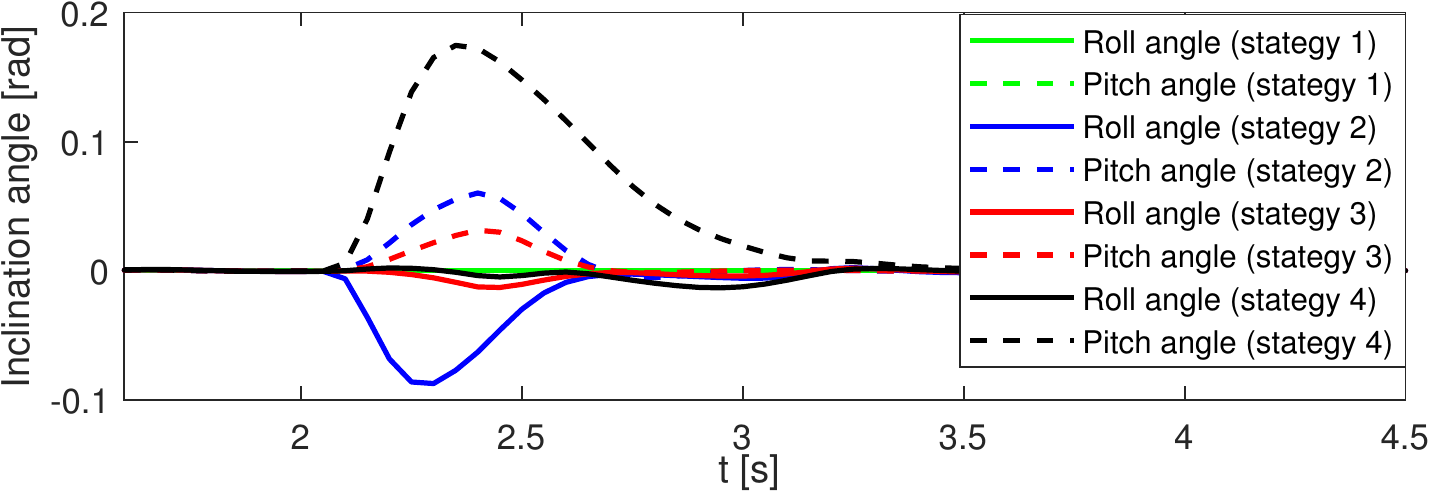}
  \caption{{Body inclination angles generated by different strategies when faced with the same push force.}}
  \label{fig:robuts23-body inclination angle}
\end{figure}

When using strategy 2, the angular momentum is modulated so as to reject external disturbances, as demonstrated in Fig. \ref{fig:robuts23-body inclination angle}. Due to the rotation of the upper-body,  the $4^{\text{th}}$ step length is modified to be 0.27 m. Besides, the $5^{\text{th}}$ step location returns to be (0.6 m,-0.635 m), which is quite close to the reference one. That is to say, compared with strategy 1, smaller parameters variation is needed to recover from the same push force, which can be seen from Fig. \ref{fig:robuts1-position} (a)$\sim$(b).

\begin{figure}
  \centering
  \includegraphics[width=\columnwidth-1mm]{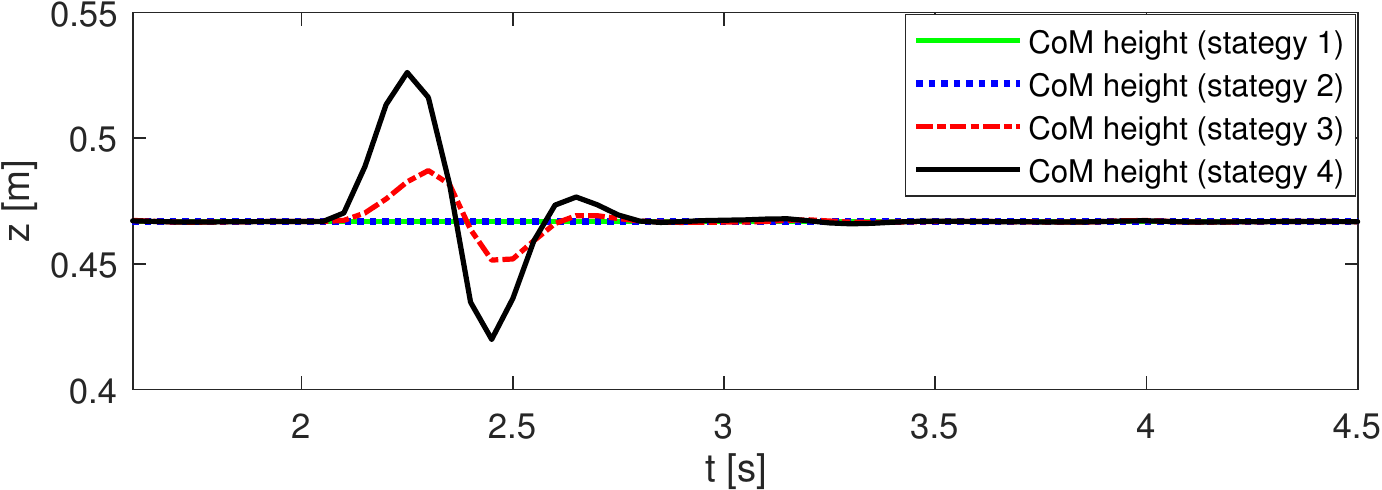}
  \caption{{CoM height trajectories generated by different strategies when faced with the same push force.}}
  \label{fig:robuts3-height}
\end{figure}

{By using strategy 3, the CoM height variation is utilized to enhance the robustness. As demonstrated in Fig. \ref{fig:robuts3-height}, when the disturbance occurs, the CoM moves up so that the excess kinetic energy partly converts into potential energy, which is similar to the result in the last section.  As a result,  the smallest variation of step length and step width is needed to reject the same external disturbance, as can be seen in Fig. \ref{fig:robuts1-position} (c). Furthermore, due to the vertical CoM variation, less body inclination is observed in Fig. \ref{fig:robuts23-body inclination angle}. Differing from \cite{lack2015integrating} and \cite{shafiee2017robust}, the optimal time-varying CoM trajectory is generated online in this work.}

As to strategy 4, as illustrated in Fig. \ref{fig:robuts1-position} (d), Fig. \ref{fig:robuts23-body inclination angle} and Fig. \ref{fig:robuts3-height}, even though the step location is not adjusted, the upper-body rotation and height variation contributes to robust walking. Under the same push force, the height variation and pitch angle variation of strategy 4 are much higher than those of other strategies. One interesting thing is that the roll angle fluctuation is dramatically suppressed. We guess it is because that {1) the height variation compensates for the lateral tracking error, 2) the lateral force is not large enough.}

\begin{table}
  \caption{{Maximal push forces the robot can reject when using four different strategies-inverted pendulum simulations.}}
  \centering
  \setlength{\tabcolsep}{1mm}
{\begin{tabular}{lcccc} \toprule
 \diagbox{Force}{Strategy} & Strategy 1 & Strategy 2 & Strategy 3& Strategy 4 \\ \midrule
 $F_\text{x}$[N]  & 139 &149 & 174& 144\\
 $F_\text{y}$[N]  & 78 & 93 & 112& 89\\
 \bottomrule
\end{tabular}}
\label{table:forces}
\end{table}

{Further analysis reveals that the integration of multiple reactive strategies contributes to stronger push recovery capability}. As listed in Table \ref{table:forces}, strategy 1 could only endure 139 N forward force $(F_{\text{x}})$ and 78 N lateral force $(F_{\text{y}})$, and strategy 2 could endure 149 N forward force and 93 N lateral force. By integrating all the three balancing strategies (actually, four strategies are integrated if taking into account the ankle strategy), the robot could recover from much larger pushes (174 N forward force and 112 N lateral force). Hence, the improved robustness against external pushes is achieved. Besides, in this case, the simulation result shows that strategy 4 contributes to the higher push recovery capability than strategy 1.

 \section{Whole-body Dynamic Simulation} \label{whole-body-simu}
\begin{figure*}
	\centering
	\includegraphics[width=140mm]{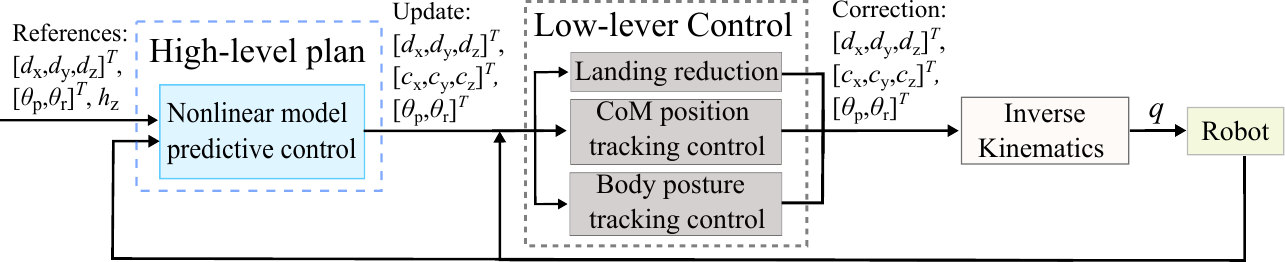}
	\caption{{Algorithm flow for robust locomotion control.}}
	\label{fig:overall-flow}
\end{figure*}

To further demonstrate the effectiveness of the proposed strategy, the whole-body dynamic simulations are conducted on the COMAN humanoid robot. In this section, the $5^{\text{th}}$ polynomial interpolation algorithm is adopted to synthesize the swing foot trajectory. The control frequency is 200Hz. To reduce the tracking error, the low-level PD controllers for CoM position tracking, upper-body inclination angle tracking are also integrated here. Besides, the landing reduction method proposed in \cite{ding2018walking} is also integrated here. {The block diagram is illustrated in Fig. \ref{fig:overall-flow}.}

\subsection{3D walking across uneven terrain} \label{3d-ode}
{In this section, the stair climbing capability is demonstrated. Since the step location is limited by the support surface, the robot would not adjust the step locations in real-time. Thus, strategy 4 is used here to generate a robust walking pattern.}

The whole-body dynamic simulations demonstrate that, by integrating the CoM height adaptation, the robot can walk stably on the 9 cm high stairs  (21\% of the total leg length), which can be seen in the attached video\cite{{video2020}}. On the contrary, when the height variation is not employed, the robot can only climb the 3 cm height stairs. 

\begin{figure*}
  \centering
  \includegraphics[width=110mm]{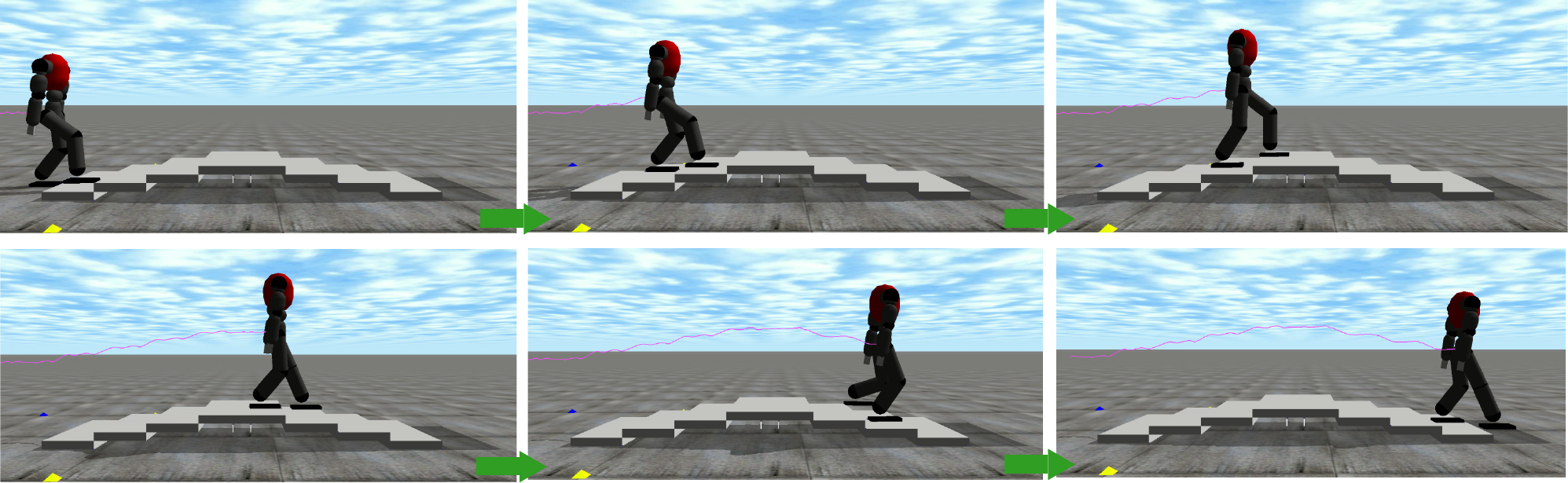}
  \caption{COMAN robot climbs stairs, the stair height is 5 cm (11.5\% of whole leg length).}
  \label{fig:3D walking-snapshot}
\end{figure*}

\begin{figure*}
  \centering
  \includegraphics[width=140mm]{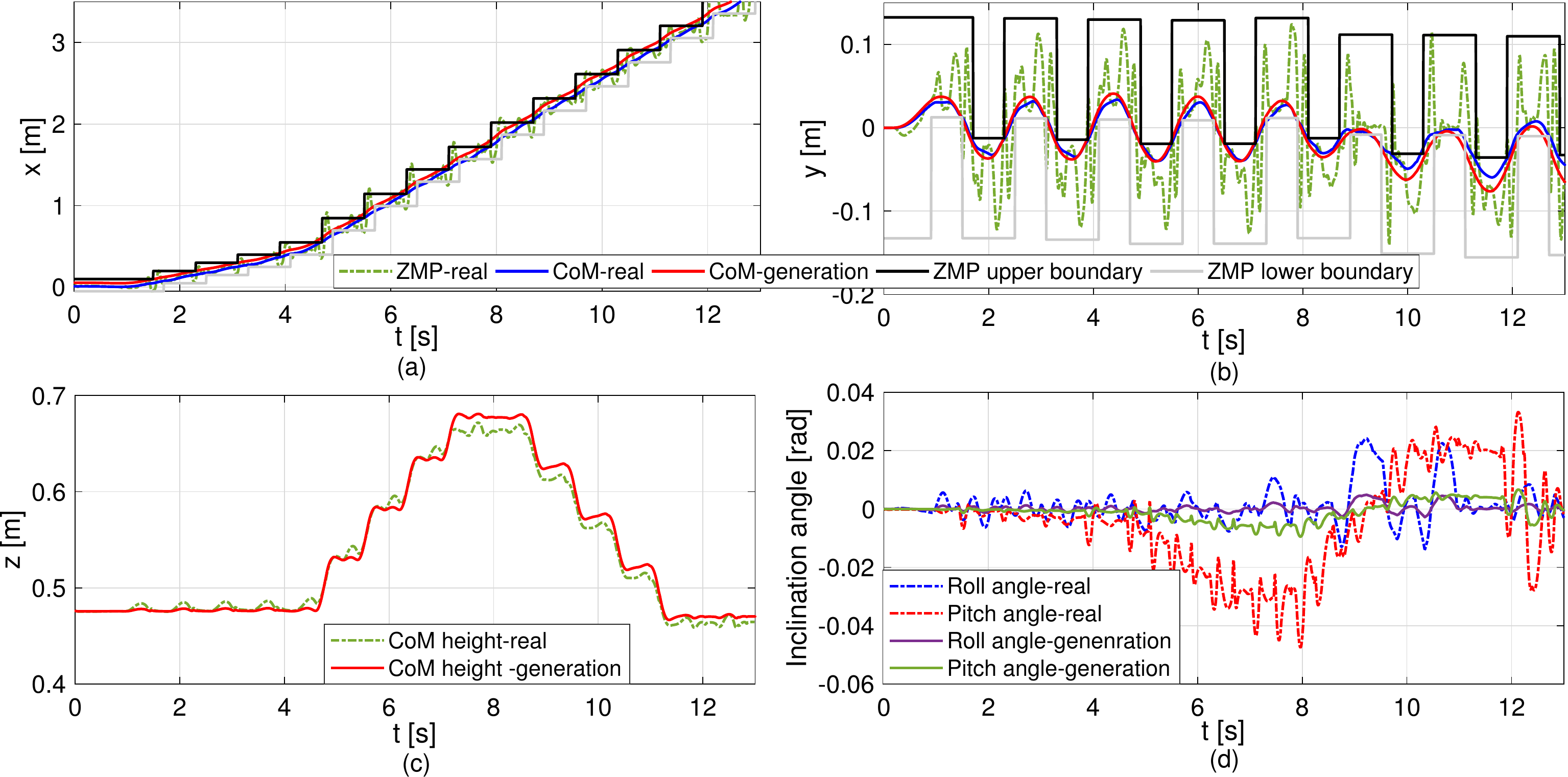}
  \caption{{Estimated and generated trajectories when climbing 5 cm stairs. (a) Forward trajectories. (b) Lateral trajectories. (c) CoM height trajectories. (d) Body inclination angles.}}
  \label{fig:3D walking-trajectory_ode}
\end{figure*}

The snapshots of stable walking across 5 cm high stairs are illustrated in Fig.  \ref{fig:3D walking-snapshot}. In this case, the robot first climbs up four stairs and then climbs down. The default step length, step width and walking cycle for walking on the flat ground are 0.15 m, 0.0145 m and 0.8 s, respectively. To avoid collision with stairs, the step length is extended to be 0.3 m when climbing stairs. Using the proposed strategy, the estimated and generated trajectories are illustrated in Fig. \ref{fig:3D walking-trajectory_ode}.

As can be seen from Fig. \ref{fig:3D walking-trajectory_ode}, the proposed framework generates the feasible 3D CoM trajectory by modulating the CoM height ((Fig. \ref{fig:3D walking-trajectory_ode} (c)) and body inclination states (Fig. \ref{fig:3D walking-trajectory_ode} (d)) as well. Further observation in Fig.\ref{fig:3D walking-trajectory_ode} (a) and Fig.\ref{fig:3D walking-trajectory_ode} (b) demonstrates that the real ZMP trajectory is restricted within the support region during most period. However, due to the landing impact, the forward ZMP trajectory easily goes beyond the foot support region during the double support phase. Besides, when climbing down the stairs, the lateral ZMP trajectory may also go beyond the stability region, especially when walking from the $12^{\text{th}}$ to the $14^{\text{th}}$ step. Nevertheless, due to the angular momentum adaptation and height variation, the robot can accomplish this walking task using the synthesized gait.

Since the step length enlarges dramatically when climbing stairs, the effect of the swing leg mass on the walking stability cannot be ignored. Consequently, the real CoM height oscillates up and down during the walking process, as can be seen in Fig.\ref{fig:3D walking-trajectory_ode} (c). {Besides, the earlier landing when stepping up the stairs and the later landing when stepping down the stairs cannot be avoid}. As a result, there exist tracking errors on the horizontal step locations and CoM position, such as the lateral movement of step locations demonstrated in Fig.\ref{fig:3D walking-trajectory_ode} (b). Furthermore, the upper-body first leans backward when stepping up the stairs and then leans forward when stepping down the stairs, as can be seen in Fig.\ref{fig:3D walking-trajectory_ode} (d).
Nevertheless, the robot still maintains balance when walking on the uneven terrain by using the proposed NMPC algorithm. Since the main task is to walking stably across the uneven terrain, we believe the tracking error is acceptable in this case.

\subsection{{Adaptive walking through the narrow space}} \label{3d-height}
{In the above sections, the default inverted pendulum height ($h_\text{z}^{\text{ref}}$) is set to be 0.467 m, and the body inclination angles ($\theta_{\text{r}}^{\text{ref}}$ and
$\theta_{\text{p}}^{\text{ref}}$) are set to be zeros during the whole walking process. However, in certain circumstances, the NMPC algorithm can be used to generate adaptive gaits by tracking the time-varying inverted pendulum height reference and body inclination angle reference, contributing to high adaptability}. One typical scenario is that the robot is required to reduce the body height and lean forward as well in order to go through a narrow passage with limited height.

\begin{figure*}
  \centering
  \includegraphics[width=120mm]{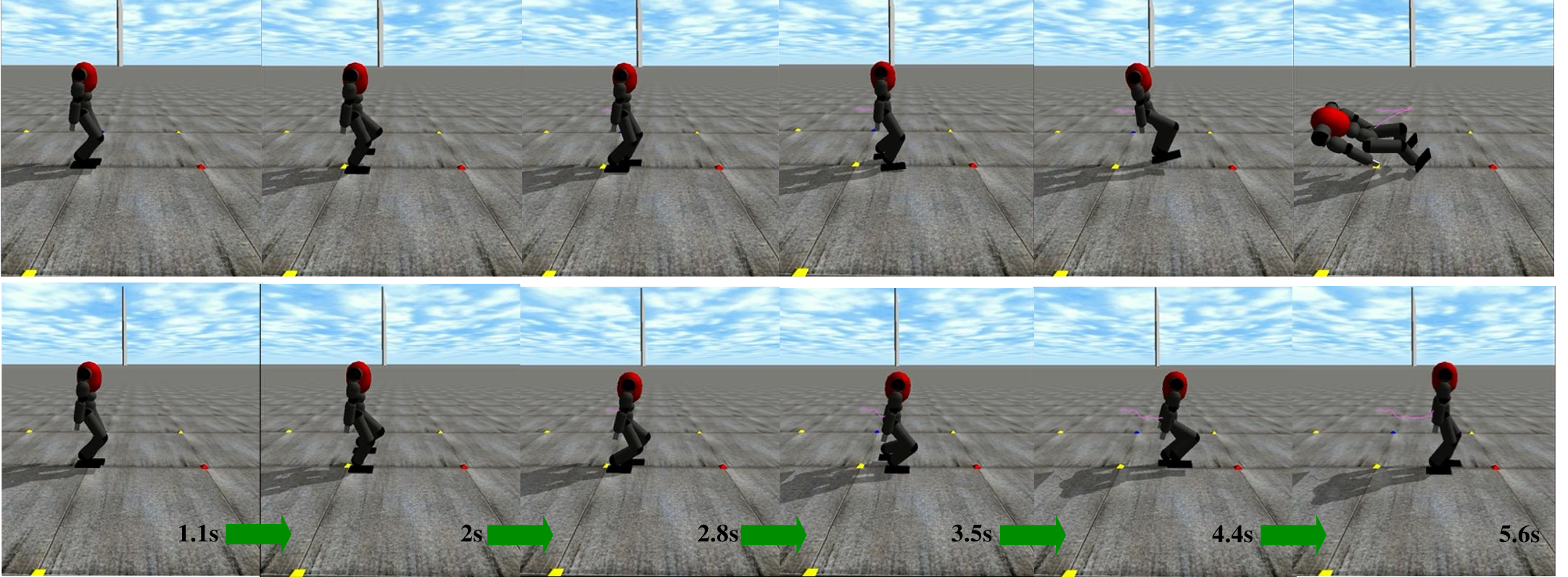}
  \caption{{COMAN robot walks through the narrow space with limited height. The top arrow demonstrates that the robot collides with the obstacle and falls down because of no usage of height variation or upper-body rotation. The below arrow demonstrates that the robot walks across the narrow crouch by reducing CoM height and rotating upper-body.}}
  \label{fig:limited_height_snap}
\end{figure*}

\begin{figure*}
  \centering
  \includegraphics[width=120mm]{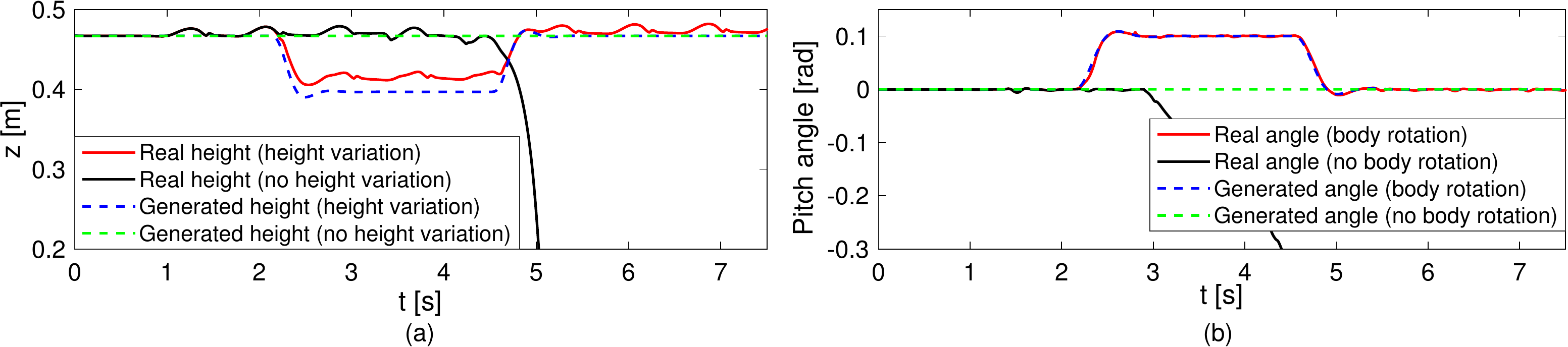}
  \caption{Generated and real trajectories for walking through the narrow space. (a) CoM height trajectories. (b) Upper-body pitch angles.}
  \label{fig:limited_height_ode}
\end{figure*}

{In the ODE environment, the obstacle is set at the 0.3 m in front of the robot and 0.75 m in height. As one solution, the robot needs to reduce the height by 5 cm while rotating the upper-body along the $y$ axis in 0.1 rad. As a comparison, no height variation or body inclination is also tested. The snapshots of walking motions when using different references are demonstrated in Fig. \ref{fig:limited_height_snap}. The height trajectories and pitch angles when using different references are illustrated in Fig. \ref{fig:limited_height_ode}.} 

{As can be seen in Fig. \ref{fig:limited_height_snap}, by tracking time-varying height reference and pitch angle reference, the NMPC algorithm generates the adaptive gait for walking through the narrow space. In particular, as illustrated in Fig. \ref{fig:limited_height_ode} (a), the robot firstly reduces the CoM height and then recovers to the normal height. The similar motion can also be found in the pitch rotation. Although there are tracking errors of the CoM height and pitch angle, the stable walking is realized. In contrast, when the height variation or upper-body rotation is not utilized, the robot collides with the obstacle and falls down.} That is to say, the hip strategy and the height variation strategy integrated into the proposed NMPC algorithm contributes to adaptive walking in real-world environments.

\subsection{Balance recovery from external pushes} \label{push-ode}
{Similar to Section \ref{robust}, we analyze the push recovery capability when using four different strategies. Firstly, the push recovery performance when the robot is stepping in place is evaluated. Then, push recovery performance when the robot is walking forward is discussed.}

\subsubsection{Balance recovery when stepping in place} \label{push-stepping}
In this case, the robot is stepping in place when the horizontal external forces are applied at the pelvis center at 3.6 s, which lasted 0.1 s. 

In this section, we discuss the balance recovery movements under forward push in detail. Under different external pushes, the generated patterns when using four different strategy combinations are demonstrated in Fig. \ref{fig:trajectory_feedbackxx}, and the physical motions can be seen in the supplementary video. 

\begin{figure*}
  \centering
  \includegraphics[width=140mm]{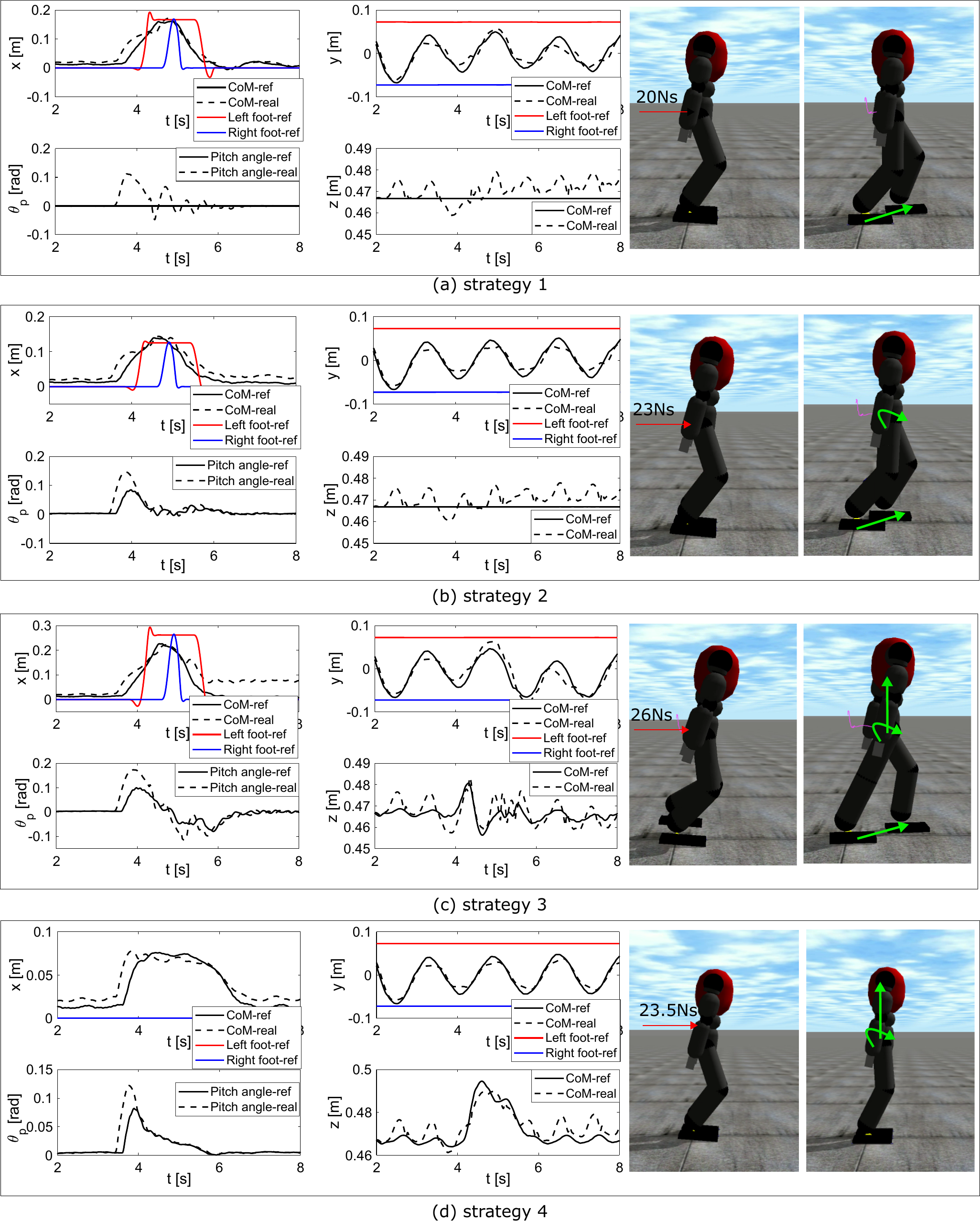}
  \caption{Reference and estimated trajectories when recovering from different levels of forward pushes. The snapshots of robot motions are demonstrated on the right side. In this case, when 20Ns external push is imposed, only the step location is modulated (strategy 1), as demonstrated by the green linear arrow in sub-figure (a). As the disturbance increases (23Ns in this case), the body rotation should be integrated (strategy 2),  as demonstrated by the green curved arrow in sub-figure (b). When 26Ns external impulse is imposed, the CoM vertical height variation  (strategy 3) is also activated,  as demonstrated by the green vertical arrow in sub-figure (c). As a comparison, merely the body rotation and vertical height variation (strategy 4) are utilized, as demonstrated in sub-figure (d).}
  \label{fig:trajectory_feedbackxx}
\end{figure*}

When the external force is not large enough, the robot can recover from the disturbance by merely making one reactive step. As demonstrated in Fig. \ref{fig:trajectory_feedbackxx} (a), the step length is modulated to be 0.18 m to extend the support region. As a result, the excess kinetic energy can be dissipated so that the body balance is maintained. Due to the coupling property, the lateral CoM is also adjusted.  

As external impulse increases, angular momentum adaptation needs to be employed. As demonstrated in Fig. \ref{fig:trajectory_feedbackxx} (b), the robot rotates the upper-body around the $y$ axis as well as modulates step location when 23 Ns external impulse is applied. Differing from Fig. \ref{fig:trajectory_feedbackxx} (a) where no reactive roll angle is generated, the maximal reference roll angle reaches 0.08 rad, as can be seen from the 'Pitch angle-ref' plot in \ref{fig:trajectory_feedbackxx} (b). Due to the utilization of upper-body rotation, even when faced with the more considerable disturbance, smaller step location modulation is needed (0.12 m in Fig. \ref{fig:trajectory_feedbackxx} (b) vs. 0.18 m Fig. \ref{fig:trajectory_feedbackxx} (a)).

When the external force goes beyond one specific limit, vertical height variation needs to be activated. As demonstrated in Fig. \ref{fig:trajectory_feedbackxx} (c), three reactive strategies are all triggered so as to maintain the balance when 26 Ns external impulse is imposed. Differing from Fig. \ref{fig:trajectory_feedbackxx} (a) and \ref{fig:trajectory_feedbackxx} (b) where no reactive height variation is employed, the CoM moves up above 0.51 m after 4 s, as can be seen from the vertical height trajectory in \ref{fig:trajectory_feedbackxx} (c). Then, after 5 s, the CoM height drops, leading to higher stability.

As a comparison, step location adjustment is not utilized for push recovery test (Strategy 4). As demonstrated in Fig. \ref{fig:trajectory_feedbackxx} (d), when the step location adjustment is not used, the robot leans forward while increasing the body height to reject the external perturbation. Due to the weight coefficients setup, the maximal reference pitch angle is 0.078 rad (the upper boundary for the pitch angle is 0.087 rad) while the maximal height reaches 0.49 m. Compared with Fig. \ref{fig:trajectory_feedbackxx} (c), it also takes a longer time period to return to the reference inclination angle and reference CoM height.

\subsubsection{Balance recovery when walking forward} \label{simulation_walking push}

{In this case, the robot is walking forward when the horizontal external forces are applied at the pelvis center at 3.6 s which lasted 0.1 s. The default step length is 0.1 m and the default step length is 0.145 m.} 

{Differing from Section \ref{ipfm_forward_walking} where the forward and lateral pushes are imposed simultaneously, in this section, we focus on the forward push recovery so as to compare the performances using different strategy combinations. Under different external pushes, the generated patterns when using four strategy combinations are illustrated in Fig. \ref{fig:forward-push_comx}- \ref{fig:forward-push_height}, and the animation can be seen in the supplementary video.} 
\begin{figure*}
  \centering
  \includegraphics[width=140mm]{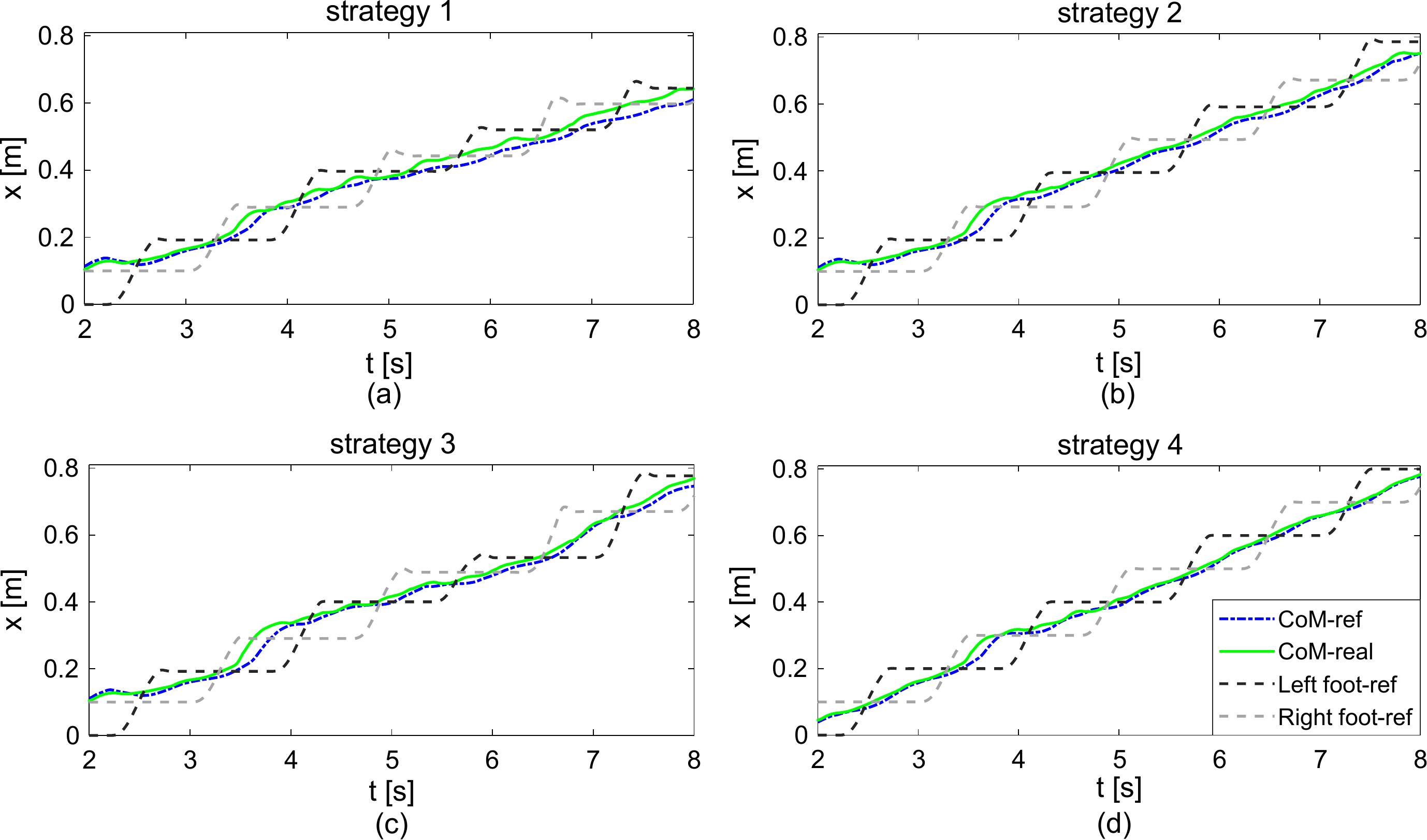}
  \caption{{Reference and estimated forward trajectories when recovering from different levels of forward pushes: (a) $F_\text{x}$ = 260 N; (b) $F_\text{x}$ = 310 N; (c) $F_\text{x}$ = 350 N; (d) $F_\text{x}$ = 295 N.}}
  \label{fig:forward-push_comx}
\end{figure*}

\begin{figure*}
  \centering
  \includegraphics[width=130mm]{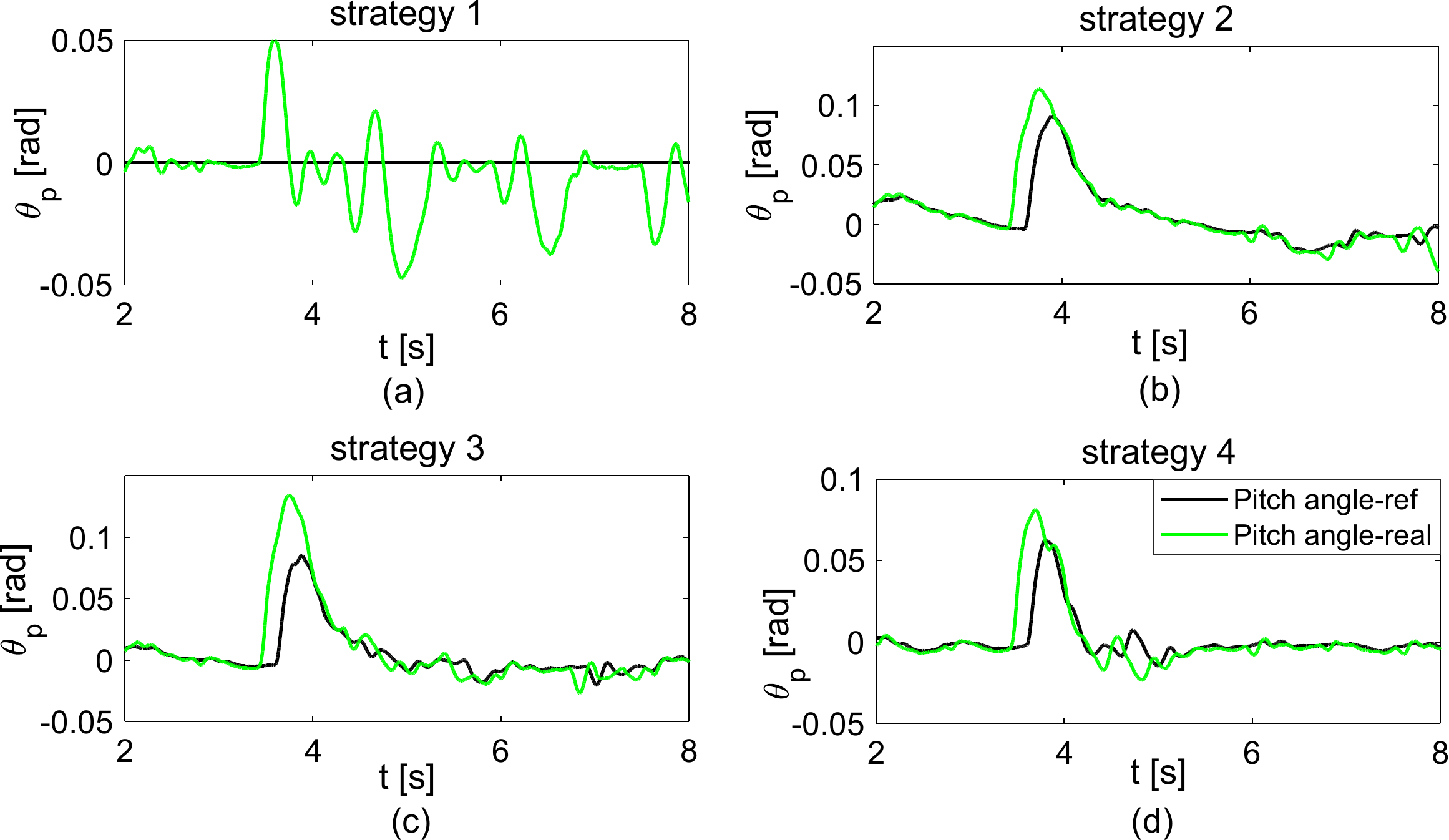}
  \caption{Reference and estimated pitch angles when recovering from different levels of forward pushes.}
  \label{fig:forward-push_body}
\end{figure*}

\begin{figure*}
  \centering
  \includegraphics[width=130mm]{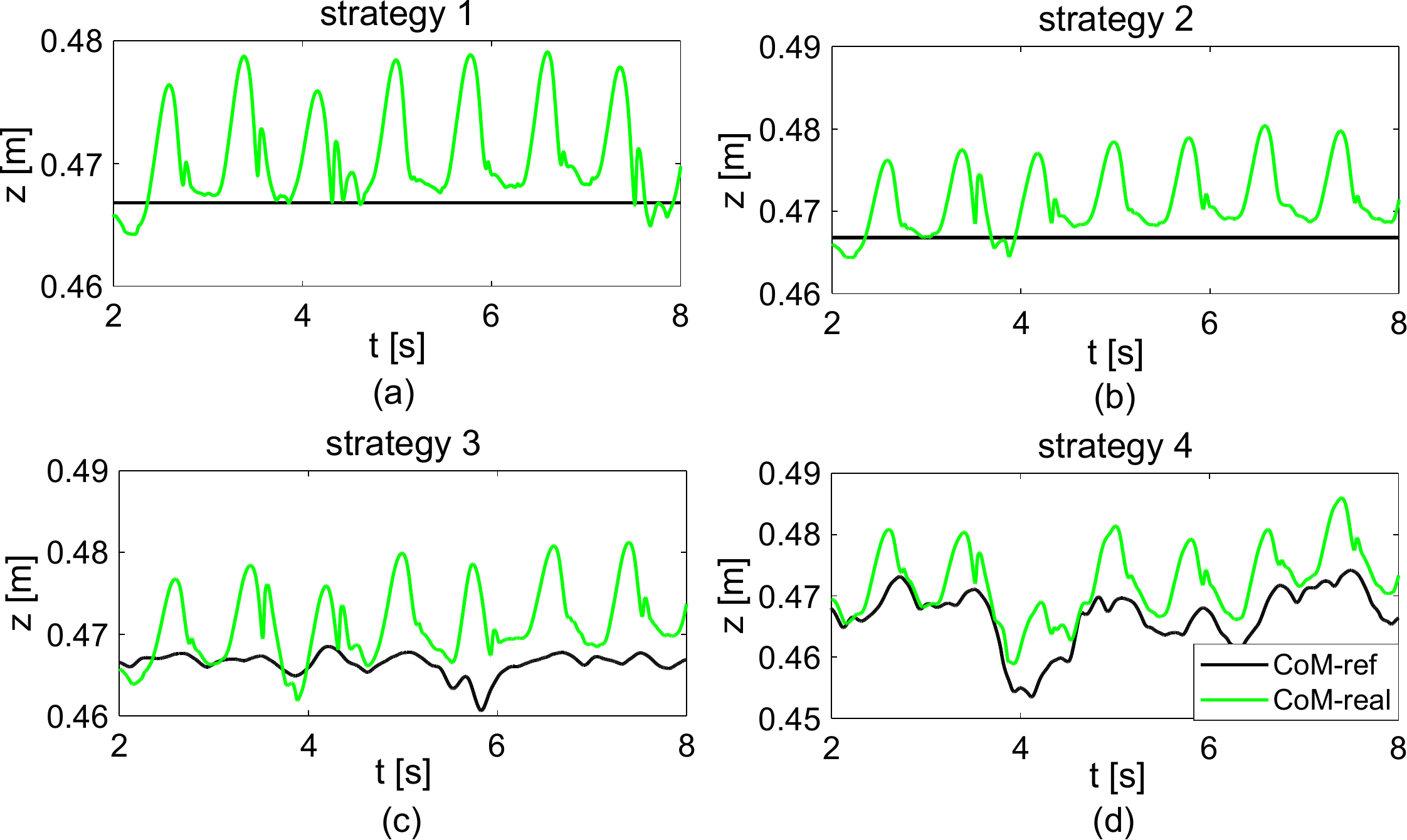}
  \caption{Reference and estimated height trajectories when recovering from different levels of forward pushes.}
  \label{fig:forward-push_height}
\end{figure*}

{As demonstrated in Fig. \ref{fig:forward-push_comx} (a), Fig. \ref{fig:forward-push_body} (a), and Fig. \ref{fig:forward-push_height} (a), where no angular momentum adaptation or height variation is employed, the robot can still maintain the balance by adjusting the step locations in following steps. As demonstrated in Fig. \ref{fig:forward-push_comx} (a), differing from Section \ref{ipfm_forward_walking} where the robot enlarges the step length immediately after the external push is imposed, the swing leg here lands the ground in a shorter length at the $4^\text{th}$ step (the step location is reduced from 0.3 m to 0.29 m). We guess it is because that, at the startup of the $4^\text{th}$ step, the estimated CoM already falls behind the step center, which can be seen in Fig. \ref{fig:forward-push_comx} (a) at about 3.3 s. Besides, there is also a time delay in state estimation. Thus, the step location is reduced at the $4^\text{th}$ step even when the forward force is imposed at 3.6 s. The similar phenomena can also be observed in other sub-figures in Fig. \ref{fig:forward-push_comx}. Then, since the CoM falls beforehand the step center at 4.1 s, the step length is enlarged to track the reference step location (0.4 m in this case). After then, since no body rotation or height variation is activated, the step locations in the following steps are also adjusted. As a result, even under the smallest external force, the tracking error of the step location when using strategy 1 becomes the largest among the four strategies.}

{As can be seen in Fig. \ref{fig:forward-push_comx}-\ref{fig:forward-push_height}, the robot can reject larger push force when more strategies is integrated. Taking strategy 3 as an example, by adjusting the step location, rotating the upper-body and changing the CoM height simultaneously, the robot can recover from the 350 N  push force.}

{In particular, when the strategy 4 is employed, only the angular momentum and CoM height are adjusted to reject the external disturbance, as demonstrated in \ref{fig:forward-push_body} (d) and Fig. \ref{fig:forward-push_height} (d). It should be mentioned that, differing from the Section \ref{ipfm_forward_walking}, the CoM height reduces dramatically when the external force is imposed. We guess that it is because that {the external force is too large. By reducing the CoM height, the moment arm could be shortened} so as to reduce the overturning moment caused by the variation of forward CoM position, considering that there is no possibility of the step location adjustment, After then, the robot increases the height and goes back to the normal height. Compared with other sub-figures in Fig. \ref{fig:forward-push_height}, there is the largest fluctuation in CoM height when using strategy 4. Nevertheless, the robot can maintain balance during the whole walking process.}

When walking forward, the maximal tolerant push forces the robot can reject are listed in Table \ref{table:forces_ode}. Again, the robot achieves the strongest recovery capability by integrating the step location adjustment, angular momentum optimization, and vertical height adaptation in one framework. That is to say, the proposed NMPC algorithm contributes to higher robustness.

\begin{table}
  \caption{Maximal tolerant push forces the robot can reject under different strategies-dynamic simulations.}
  \centering
  	\setlength{\tabcolsep}{1mm}
{\begin{tabular}{lcccc} \toprule
 \diagbox{Param.}{Strategy} & Strategy 1 & Strategy 2 & Strategy 3 & Strategy 4 \\ \midrule
 $F_\text{x}$[N]  &{273}&{335}&{376}&{295}\\
 $F_\text{y}$[N]  &{160}&{185}&{255}&{166}\\
 \bottomrule
\end{tabular}}
\label{table:forces_ode}
\end{table}

\subsection{Computation Efficiency}\label{compute-effi}
{ In this paper, the SQP algorithm (\ref{eq:sqp}) is used to solve the QCQP (\ref{eq:qcqp}). Differing from \cite{van2017real} where the $ N_s$ was set to be 2 by hand, we propose to use the following termination function:}
\begin{equation} \label{eq:ternimation condtion}
\begin{aligned}
{\min (\mathbf{F}^m)} & {\leq} {\varepsilon}  \quad  where \quad  {\mathbf{F}^m_{(n)}=\max(\mathbf{\Delta}_{{n_k}(n)}),}\\\
n_{k} &> N_s.  \\
\end{aligned}
\end{equation}
\normalsize
where $\mathbf{\Delta}_{{n_k}}$, consisting of the increments of 3D CoM jerk, 2D angular jerk and 3D step location, is computed by the QP solver, $\mathbf{F}^m \in \Re^{8}$, $n_k$ is the loop count of QP solver.

That is to say, there are two ways to terminate the SQP loop. {The first way is that we firstly judge the maximal absolute value of each state channel, including 3D CoM, 2D body inclination angle, and 3D step location, over the prediction horizon. Then, we need to compare the minimum among the maximal values with the $\varepsilon$}. The second way is the judge if the loop number goes beyond the limitation. Therefore, before using the condition expressed in \eqref{eq:ternimation condtion}, the proper $ \varepsilon$ and $N^{\max}$ should be determined so as to balance the trade-off between computation cost and optimization performance.

\begin{figure*}
  \centering
  \includegraphics[width=140mm]{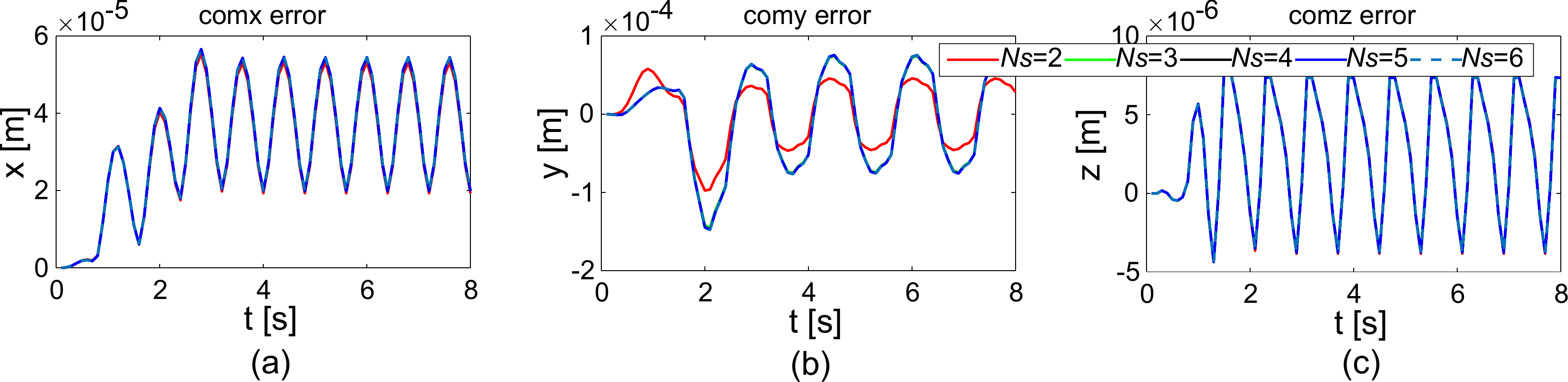}
  \caption{The CoM errors when using different $N_s$: the benchmark is the generated CoM trajectory when $N_s=1$. }
  \label{tracking_error_text}
\end{figure*}

To determine the  $ \varepsilon$ and $N^{\max}$, the gait pattern generation tests are conducted by using the QP solver provided in the C++ optimization library \textit{QuadProg++} to solve the NMPC problem. In this case, the robot walks with constant step length (0.1 m), step width (0.145 m), and step duration (0.8 s). The sampling time $dt$ is 0.1 s and the time duration of the prediction horizon is 1.0 s.  When using different $N_s$, the optimization performance of CoM generation is illustrated in Fig. \ref{tracking_error_text} and the statistical result about the minimal $\varepsilon$s and average time costs (on a 3.3 GHz quad-core CPU) are listed in Table \ref{computation_cost}.

\begin{table}
  \caption{Average time costs and minimal thresholds w.r.t the loop number.}
    \centering
  \setlength{\tabcolsep}{0.4mm}
{\begin{tabular}{lcccccc} \toprule
 \diagbox{Param.}{$N_s$} &  1 & 2 & 3 & 4 & 5 & 6 \\ \midrule
 $\varepsilon$  &1$\times10^{-2}$&5$\times10^{-7}$&5$\times10^{-8}$&4$\times10^{-9}$&7$\times10^{-10}$&5$\times10^{-11}$\\
 time cost[ms]  &{1.52}&{2.64}&{3.56}&{4.56}&{5.52}&{6.57}\\
 \bottomrule
\end{tabular}}
\label{computation_cost}
\end{table}

{As can be seen in Fig. \ref{tracking_error_text}, especially in the Fig. \ref{tracking_error_text}(b), the CoM error decreases as the $N_s$ increases.} However, when $N_s$ becomes larger than 3, the effect is dramatically weakened. The similar rule can be found in the $\varepsilon$, which is listed in Table \ref{computation_cost}. On the other side, the (average) time cost increases linearly with the increase of $N_s$. Thus, using a too large loop number which is bigger than 3 is not advised. As a result, for the real application, we can choose the parameters combination where $\varepsilon=5\times10^{-8} $ and $N_s=3$ for QCQP solution.

\section{Conclusion} \label{conclusion}
In this paper, we propose a versatile framework for robust locomotion. The proposed framework is formulated as a QCQP and solved via the off-the-shelf SQP technique.

By using the NIPFM, which can take into consideration the CoM height variation and angular momentum change, the ZMP constraints are formulated in a quadratic form. Through combining step movement limitations and other feasibility constraints, robust bipedal walking is realized with the capabilities of ZMP manipulation, stepping location adjustment, CoM height variation and upper-body rotation. In addition, the proposed framework can generate versatile walking patterns by customizing the reference CoM height trajectory or body inclination angle. Simulation studies demonstrate that the robot is able to achieve higher adaptability under multiple real-world scenarios. 

We believe a promising future about this framework since it can be used to generate the adaptive walking patterns and to maintain balance as well for humanoids in the real environment. At present, we tune the penalties in the objective function by hand, which is tedious. Thus, studying the priority of different strategies under realistic scenarios can be the next focus.

\appendix
 
 \section{Parameters for Objective Function} \label{objective function parameters}
 For the objective function defined in (\ref{eq:obj_Model}), using (\ref{eq:MPC_state_re}), the $\mathbf{G}$, $\mathbf{g}$ and state $\bm{\mathcal{X}}$ in (\ref{eq:qcqp}) are expressed as follows:
\begin{equation} \label{eq:obj_para}
\centering
\begin{aligned}
\mathbf{G} = \textit{diag}(\vartheta_{_{\mathbf{C}_{\text{x}}}}, \vartheta_{_{\mathbf{C}_{\text{y}}}}, \vartheta_{_{\mathbf{C}_{\text{z}}}}, \vartheta_{_{\mathbf{\mathbf{\Theta}}_{\text{r}}}}, \vartheta_{_{\mathbf{\mathbf{\Theta}}_{\text{p}}}}, \phi_{_{\mathbf{D}_{\text{x}}}}, \phi_{_{\mathbf{D}_{\text{y}}}}, \phi_{_{\mathbf{D}_{\text{z}}}}), \qquad \\
\vartheta_{_{\mathbf{X}}} = \frac{\gamma_{_{\mathbf{X}}}}{2} \mathbf{I}_{N_h \times N_h} +\frac{\alpha_{_{\mathbf{X}}}}{2} \mathbf{P}_{\text{vu}}^T \mathbf{P}_{\text{vu}}  + \frac{\beta_{_{\mathbf{X}}}}{2} \mathbf{P}_{\text{pu}}^T \mathbf{P}_{\text{pu}}, \qquad \quad \\
\phi_{_{\mathbf{U}}} = \frac{\delta_{_{\mathbf{U}}}}{2} \mathbf{I}_{N_f \times N_f}, \qquad \qquad \qquad \qquad \\
\mathbf{g}=
\left[ \! \! 
\begin{array} {c}
(\alpha_{_{\mathbf{C}_{\text{x}}}} \mathbf{P}_{\text{vu}}^T \mathbf{P}_{\text{vs}} + \beta_{_{\mathbf{C}_{\text{x}}}} \mathbf{P}_{\text{pu}}^T \mathbf{P}_{\text{ps}}) \hat{c}_{\text{x}(k)}- \beta_{_{\mathbf{C}_{\text{x}}}} \mathbf{P}_{\text{pu}}^T \mathbf{C}_{\text{x}(k)}^{\text{ref}} \\
(\alpha_{_{\mathbf{C}_{\text{y}}}} \mathbf{P}_{\text{vu}}^T \mathbf{P}_{\text{vs}} + \beta_{_{\mathbf{C}_{\text{y}}}} \mathbf{P}_{\text{pu}}^T \mathbf{P}_{\text{ps}}) \hat{c}_{\text{y}(k)} - \beta_{_{\mathbf{C}_{\text{y}}}} \mathbf{P}_{\text{pu}}^T \mathbf{C}_{\text{y}(k)}^{\text{ref}} \\
(\alpha_{_{\mathbf{C}_{\text{z}}}} \mathbf{P}_{\text{vu}}^T \mathbf{P}_{\text{vs}} + \beta_{_{\mathbf{C}_{\text{z}}}} \mathbf{P}_{\text{pu}}^T \mathbf{P}_{\text{ps}}) \hat{c}_{\text{z}(k)} - \beta_{_{\mathbf{C}_{\text{z}}}} \mathbf{P}_{\text{pu}}^T \mathbf{C}_{\text{z}(k)}^{\text{ref}} \\ 
(\alpha_{_{\mathbf{\Theta}_{\text{r}}}} \mathbf{P}_{\text{vu}}^T \mathbf{P}_{\text{vs}} + \beta_{_{\mathbf{\Theta}_{\text{r}}}} \mathbf{P}_{\text{pu}}^T \mathbf{P}_{\text{ps}}) \hat{\theta}_{\text{r}(k)} - \beta_{_{\mathbf{\Theta}_{\text{r}}}} \mathbf{P}_{\text{pu}}^T \mathbf{\Theta}_{\text{r}(k)}^{\text{ref}} \\
(\alpha_{_{\mathbf{\Theta}_{\text{p}}}} \mathbf{P}_{\text{vu}}^T \mathbf{P}_{\text{vs}} + \beta_{_{\mathbf{\Theta}_{\text{p}}}} \mathbf{P}_{\text{pu}}^T \mathbf{P}_{\text{ps}}) \hat{\theta}_{\text{p}(k)} - \beta_{_{\mathbf{\Theta}_{\text{p}}}} \mathbf{P}_{\text{pu}}^T \mathbf{\Theta}_{\text{p}(k)}^{\text{ref}} \\
-\delta_{_{\mathbf{D}_{\text{x}}}} \mathbf{D}_{\text{x}(k)}^{\text{ref}} \\ 
-\delta_{_{\mathbf{D}_{\text{y}}}} \mathbf{D}_{\text{y}(k)}^{\text{ref}} \\
-\delta_{_{\mathbf{D}_{\text{z}}}} \mathbf{D}_{\text{z}(k)}^{\text{ref}} \\
\end{array}
\! \! \right], \\
\bm{\mathcal{X}}_{(k)} = [\dddot{\mathbf{C}}_{\text{x}(k)};\dddot{\mathbf{C}}_{\text{y}(k)};\dddot{\mathbf{C}}_{\text{z}(k)};\dddot{\mathbf{\Theta}}_{\text{r}(k)};\dddot{\mathbf{\Theta}}_{\text{p}(k)};\mathbf{D}_{\text{x}(k)};\mathbf{D}_{\text{y}(k)};\mathbf{D}_{\text{z}(k)}],
\end{aligned}
\end{equation}
\normalsize
where, \textit{diag}() is the function that produces a diagonal matrix with the given parameters in the diagonal positions, $\mathbf{X} \in \{\mathbf{C}_{\text{x}}, \mathbf{C}_{\text{y}}, \mathbf{C}_{\text{z}}, \mathbf{\Theta}_{\text{r}}, \mathbf{\Theta}_{\text{p}} \}$, $\mathbf{U} \in \{\mathbf{D}_{\text{x}}, \mathbf{D}_{\text{y}}, \mathbf{D}_{\text{z}}\}$. 


\section{Quadratic Form of Feasibility Constraints}  \label{appendixb}
All the constraints introduced in Section \ref{constraints} can be formulated in quadratic form as expressed in (\ref{eq:qcqp}). In this section, we present more details about how to express the ZMP constraints.

 At the $k^{\text{th}}$  sampling time, defining the selection matrices, the predictive CoM jerk and step locations can be expressed as follows:
\begin{equation} \label{eq:state_selection}
\begin{aligned}
\mathbf{U}_{(k)} & = \mathbf{S}_{_\mathbf{U}} \bm{\mathcal{X}}_{(k)}, && \mathbf{S}_{\text{u}} \in \Re^{N_f \times N_t}, \\ 
\dddot{\mathbf{X}}_{(k)} & = \mathbf{S}_{\mathbf{x}} \bm{\mathcal{X}}_{(k)}, && \mathbf{S}_{\text{x}} \in \Re^{N_h \times N_t},\\
\dddot{x}_{(k+j)} & = \mathbf{S}_j \dddot{\mathbf{X}}_{(k)},  && \mathbf{S}_j \in \Re^{1 \times N_h}, j \!\in \!\{1,...,N_h \}.
\end{aligned}
\end{equation}
\normalsize

Besides, the step locations over the prediction horizon are generated by using the separated locations corresponding to the following walking cycles, we have following relationship (taking the step location along the $x$- axis for instance):
\begin{equation} \label{eq:footstep_selection}
\begin{aligned}
\bar{\mathbf{D}}_{\text{x}(k)}&= \mathbf{e}_{\text{c}(k)} \hat{d}_{\text{x}(k)} + \mathbf{E}_{\text{c}(k)} \mathbf{D}_{\text{x}(k)}\\
\end{aligned}
\end{equation}
\normalsize
where, $\bar{\mathbf{D}}_{\text{x}(k)}=[d_{\text{x}(k+1)}^{\text{ref}},...,d_{\text{x}(k+N_h)}^{\text{ref}}]^T$ consists of the  step locations  over  the  prediction  horizon, $\hat{d}_{\text{x}(k)}$ denotes position of the current supporting foot, the $\mathbf{e}_{\text{c}(k)}$ and $\mathbf{E}_{\text{c}(k)}$ are mapping matrix, with more details can be found in \cite{diedam2008online}.  

And then, the position of supporting foot is determined by
\begin{equation} \label{eq:supprt_footstep_selection}
\begin{aligned}
d_{\text{x}{(k+j)}} & = \mathbf{S}_j \bar{\mathbf{D}}_{\text{x}(k)},  && j \!\in \!\{1,...,N_h \}.
\end{aligned}
\end{equation}
\normalsize

\textit{quadratic form of ZMP constraints}: To be brief, we only discuss the upper boundary. Taking the motion along the $x$- axis for instance, substitute (\ref{eq:IPFMmodel1}) into (\ref{eq:zmp_constrain}), we have:
 \begin{equation} \label{eq:zmp_constrain_express}
 \begin{aligned}
&(c_{\text{x}(k+j)}-d_{\text{x}(k+j)}-p^{\max}_{\text{x}})(g+\ddot{c}_{\text{z}(k+j)}) \\ 
&- (c_{\text{z}(k+j)}  - d_{\text{z}(k+j)}) \ddot{c}_{x(k+j)} -I_{\text{y}} \ddot{\theta}_{\text{p}(k+j)}/m  \leq 0
 \end{aligned}
 \end{equation}
 \normalsize

 Then, by substituting (\ref{eq:MPC_state_re}), (\ref{eq:state_selection}) and (\ref{eq:supprt_footstep_selection}) into (\ref{eq:zmp_constrain_express}) and collecting terms, the quadratic form of ZMP constraints is:
 \begin{equation} \label{eq:constrain_V}
 \begin{aligned}
 \mathbf{V}_{p_{\text{x}(j)}} &=m( \mathbf{S}_{\mathbf{c}_\text{x}}^T \mathbf{P}_{\text{pu}}^T \mathbf{S}_j^T \mathbf{S}_j \mathbf{P}_{\text{au}} \mathbf{S}_{\mathbf{c}_\text{z}} - \mathbf{S}_{\mathbf{c}_\text{x}}^T \mathbf{P}_{\text{au}}^T \mathbf{S}_j^T \mathbf{S}_j \mathbf{P}_{\text{pu}} \mathbf{S}_{\mathbf{c}_\text{z}} \\
 &- \mathbf{S}_{_{\mathbf{D}_\text{x}}}^T \mathbf{E}_{\text{c}(k)}^T \mathbf{S}_j^T \mathbf{S}_j \mathbf{P}_{\text{au}} \mathbf{S}_{\mathbf{c}_\text{z}}),
 \end{aligned}
 \end{equation}
 \normalsize
 
 
 \begin{equation} \label{eq:constrain_v1}
 \begin{aligned}
 \mathbf{v}_{p_{\text{x}(j)}} &= m (\hat {c}_{\text{x}(k)}^T \mathbf{P}_{\text{ps}}^T \mathbf{S}_j^T \mathbf{S}_j \mathbf{P}_{\text{au}} \mathbf{S}_{\mathbf{c}_	\text{z}} + \hat {c}_{\text{z}(k)}^T \mathbf{P}_{\text{as}}^T \mathbf{S}_j^T \mathbf{S}_j \mathbf{P}_{\text{pu}} \mathbf{S}_{\mathbf{c}_\text{x}} \\
 &+ g \mathbf{S}_j \mathbf{P}_{\text{pu}} \mathbf{S}_{\mathbf{c}_\text{x}} - (\hat {c}_{\text{z}(k)}^T \mathbf{P}_{\text{ps}}^T \mathbf{S}_j^T \mathbf{S}_j \mathbf{P}_{\text{au}} \mathbf{S}_{\mathbf{c}_\text{x}} \\
  &+ \hat {c}_{\text{x}(k)}^T \mathbf{P}_{\text{as}}^T \mathbf{S}_j^T \mathbf{S}_j \mathbf{P}_{\text{pu}} \mathbf{S}_{\mathbf{c}_	\text{z}}) + d_{\text{z}{(k+j)}} \mathbf{S} _j \mathbf{P}_{\text{au}} \mathbf{S}_{\mathbf{c}_\text{x}} \\
 &- (\hat {c}_{\text{z}(k)}^T \mathbf{P}_{\text{as}}^T \mathbf{S}_j^T \mathbf{S}_j \mathbf{E}_{\text{c}(k)} \mathbf{S}_{_{\mathbf{D}_\text{x}}}+ \hat{d}_{\text{x}(k)} \mathbf{e}_{\text{c}(k)}^T  \mathbf{S}_j^T \mathbf{S}_j \mathbf{P}_{\text{au}} \mathbf{S}_{\mathbf{c}_\text{z}}) \\
 &- g \mathbf{S}_j \mathbf{E}_{\text{c}(k)} \mathbf{S}_{_{\mathbf{D}_\text{x}}} - p^{\max}_{\text{x}} \mathbf{S}_j \mathbf{P}_{\text{au}} \mathbf{S}_{\mathbf{c}_\text{z}})^T - (I_{\text{y}} \mathbf{S}_j \mathbf{P}_{\text{au}} \mathbf{S}_{_{\mathbf{\Theta}_{\text{p}}}})^T,
 \end{aligned}
 \end{equation}
 \normalsize
 
 \begin{equation} \label{eq:constrain_cc}
 \begin{aligned}
 \sigma_{p_{\text{x}(j)}} &= m (\hat {c}_{\text{x}(k)}^T \mathbf{P}_{\text{ps}}^T \mathbf{S}_j^T  \mathbf{S}_j \mathbf{P}_{\text{as}} \hat {c}_{\text{z}(k)}  + g \mathbf{S}_j \mathbf{P}_{\text{ps}} \hat{c}_{\text{x}(k)} 
 \\
 &- \hat{c}_{\text{x}(k)}^T \mathbf{P}_{\text{as}}^T \mathbf{S}_j^T \mathbf{S}_j  \mathbf{P}_{\text{ps}} \hat{c}_{\text{z}(k)} + \hat {c}_{\text{x}(k)}^T \mathbf{P}_{\text{as}}^T \mathbf{S}_j^T d_{\text{z}{(k+j)}}
 \\
 &- \hat{d}_{\text{x}(k)} \mathbf{e}_{\text{c}(k)}^T  \mathbf{S}_j^T   \mathbf{S}_j \mathbf{P}_{\text{as}} \hat {c}_{\text{z}(k)}\\
 &- g \mathbf{S}_j \mathbf{e}_{\text{c}(k)} \hat{d}_{\text{x}(k)} - p^{\max}_{\text{x}} \mathbf{S}_j \mathbf{P}_{\text{as}} \hat {c}_{\text{z}(k)} - g p^{\max}_{\text{x}}) \\
 &- I_{\text{y}} \mathbf{S}_j \mathbf{P}_{\text{as}} \hat{\theta}_{\text{p}(k)}.
 \end{aligned}
 \end{equation}
 \normalsize

\section{Parameters Setup for Simulations}
The essential parameters for the NIPFM simulations and whole-body dynamic simulations can be found in Table \ref{table:Feasibility constraints}. 

\begin{table}[ht]
  \centering
  \renewcommand\arraystretch{1.1}
  \caption{Boundary conditions for feasibility constraints}
  \label{table:Feasibility constraints}
  \begin{tabular}{cc|cc}
	\toprule
		\multicolumn{2}{c|}{$\text{ ZMP constraints}$}  &{$\dot{d}_\text{y}^{\min}[\mathrm{m \! \cdot \!  s^{-1}}]$}&{-1}\\ 
		\cline{1-2}
		{$p_\text{x}^{\min}[\mathrm{m}[$}&{-0.03}&{$\dot{d}_\text{y}^{\max}[\mathrm{m \! \cdot\!  s^{-1}}]$}&{1}\\
		\cline{3-4} 
		{$p_\text{x}^{\max}[\mathrm{m}]$}&{0.07}&\multicolumn{2}{c}{$\text{CoM motion constraints}$}\\
		\cline{3-4} 
		{$p_\text{y}^{\min}[\mathrm{m}]$}&{-0.05}&{$h^{\min}[\mathrm{m}]$}&{-0.15}\\
		{$p_\text{y}^{\max}[\mathrm{m}]$}&{0.05}&{$h^{\max}[\mathrm{m}]$}&{0.1}\\
		\cline{1-4} 
		\multicolumn{2}{c|}{$\text{Step location constraints}$}&\multicolumn{2}{c}{$\text{Body inclination Constraints}$}\\
		\cline{1-4} 
		{$d_\text{x}^{\min}[\mathrm{m}]$}&{-0.1}&{$\theta_{\text{r}}^{\min}[\text{rad}]$}&{-0.087}\\
		{$d_\text{x}^{\max}[\mathrm{m}]$}&{0.3}&{$\theta_{\text{r}}^{\max}[\text{rad}]$}&{0.175}\\
		{$d_\text{y}^{\min}[\mathrm{m}]$}&{0.11}&{$\theta_{\text{p}}^{\min}(-\theta_{\text{p}}^{\max})[\text{rad}]$}&{-0.175}\\
		 \cline{3-4} 
		{$d_\text{y}^{\max}[\mathrm{m}]$}&{0.2}&\multicolumn{2}{c}{$\text{Torque output constraints}$}\\
		 \cline{3-4} 		
		{$\dot{d}_\text{x}^{\min}[\mathrm{m \! \cdot\!  s^{-1}}]$}&{-1}&{$\tau_{\text{r}}^{\min}(-\tau_{\text{r}}^{\max})[\mathrm{N \! \cdot\!  m}]$}&{-80}\\
		{$\dot{d}_\text{x}^{\max}[\mathrm{m \! \cdot\!  s^{-1}}]$}&{3}&{$\tau_{\text{p}}^{\min}(-\tau_{\text{p}}^{\max})[\mathrm{N \! \cdot\!  m}]$}&{-80}\\
	\bottomrule  
  \end{tabular}
\end{table}

\section*{Acknowledgment} \label{ack}
This work is supported by National Natural Science Foundation of China (Grant No. 51675385) and European Union's Horizon 2020 robotics program CogIMon (ICT-23-2014, 644727). 

\bibliography{main.bib}
\bibliographystyle{myIEEEtran}
\end{document}